\documentclass[runningheads]{llncs}
\usepackage[T1]{fontenc}
\usepackage{graphicx}
\usepackage{booktabs}
\usepackage[misc]{ifsym}

\usepackage{mwe}

\usepackage{hyperref}
\usepackage{wrapfig}
\usepackage{xcolor}
\usepackage{todonotes}
\usepackage{multirow}
\usepackage{multicol}
\usepackage{amsmath}
\usepackage{amssymb}
\usepackage{amsfonts}
\usepackage{bm}
\usepackage{placeins}

\newcommand{\data}{P2S\xspace}
\newcommand{\datafull}{Production Press Sensor Data\xspace}
\newcommand{\methodname}{RioT}
\newcommand{\methodnames}{RioT's\xspace}
\newcommand{\rrt}{\methodname\xspace}
\newcommand{\rrtf}{\methodname\textsubscript{freq}\xspace}
\newcommand{\rrts}{\methodname\textsubscript{sp}\xspace}
\newcommand{\rrtfs}{\methodname\textsubscript{freq,sp}\xspace}
\newcommand{\rrtfull}{Right on Time\xspace}

\newcommand{\codenote}{\footnote{\url{https://github.com/ml-research/RioT}}}

\newcommand\blfootnote[1]{%
  \begingroup
  \begin{NoHyper}%
  \renewcommand\thefootnote{}\footnote{\color{black}#1}%
  \addtocounter{footnote}{-1}%
  \end{NoHyper}%
  \endgroup
}

\begin{document}

\title{Right on Time: Revising Time Series Models by Constraining their Explanations}

\titlerunning{Revising Time Series Models by Constraining their Explanations}

\author{Maurice Kraus\inst{1\phantom{i}*}\textsuperscript{(\Letter)} \and
David Steinmann\inst{1, 2 \phantom{i}*} \and
Antonia Wüst\inst{1} \and
Andre Kokozinski\inst{5} \and
Kristian Kersting\inst{1, 2, 3, 4}
}

\institute{Artificial Intelligence and Machine Learning Group, TU Darmstadt
\email{\{maurice.kraus,david.steinmann\}@cs.tu-darmstadt.de}
\and
Hessian Center for Artificial Intelligence (hessian.AI), Darmstadt \and
Centre for Cognitive Science, TU Darmstadt \and
German Center for Artificial Intelligence (DFKI) \and
Institute for Production Engineering and Forming Machines, TU Darmstadt
}
\tocauthor{Maurice Kraus, David Steinmann, Antonia Wüst, Andre Kokozinski, Kristian Kersting}
\toctitle{Right on Time: Revising Time Series Models by Constraining their Explanations}
\authorrunning{Kraus et al.}

\maketitle              %

\begin{abstract}
Deep time series models often suffer from reliability issues due to their tendency to rely on spurious correlations, leading to incorrect predictions. To mitigate such shortcuts and prevent "Clever-Hans" moments in time series models, we introduce \rrtfull (\rrt), a novel method that enables interacting with model explanations across both the \textit{time} and \textit{frequency} domains. By incorporating feedback on explanations in both domains, \rrt constrains the model, steering it away from annotated spurious correlations. This dual-domain interaction strategy is crucial for effectively addressing shortcuts in time series datasets. We empirically demonstrate the effectiveness of \rrt in guiding models toward more reliable decision-making across popular time series classification and forecasting datasets, as well as our newly recorded dataset with naturally occuring shortcuts, \data, collected from a real mechanical production line.\blfootnote{$^{\ast}$These authors share equal contribution.\\\\Accepted for publication at ECML PKDD 2025  

}
\end{abstract}

\section{Introduction}
\begin{figure}[t]
    \centering
    \includegraphics[width=0.95\linewidth]{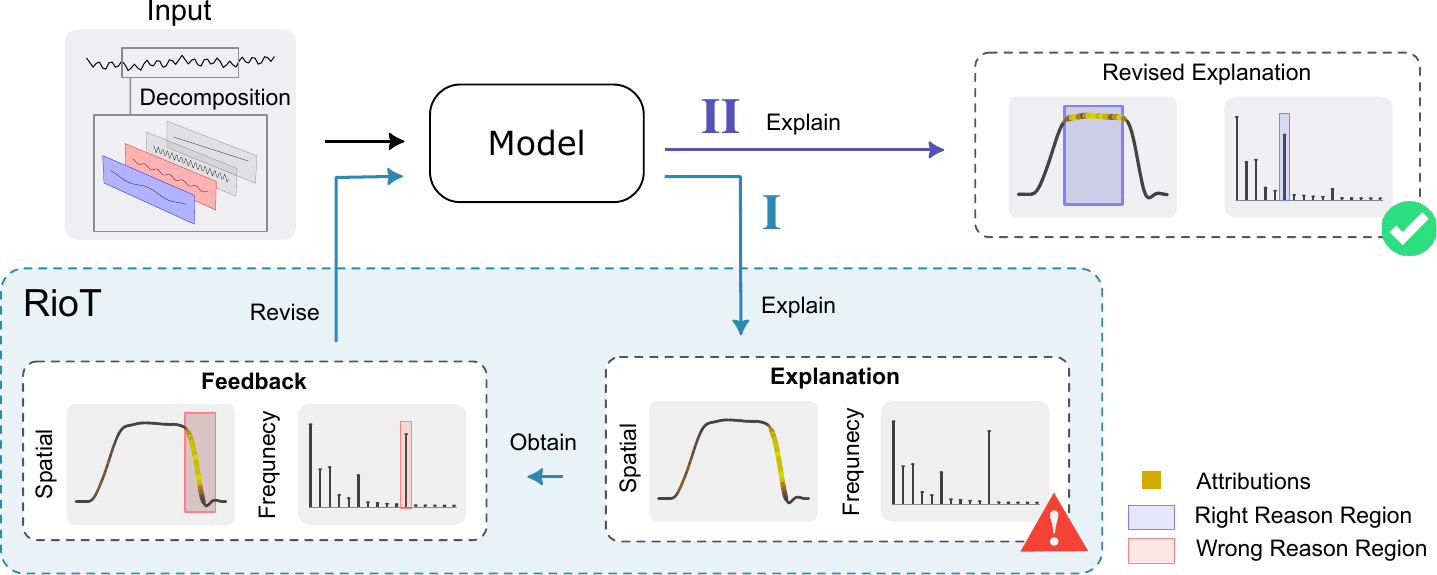}
    \caption{
    \textbf{Explanations reveal that the model relies on spurious factors in the input (\textcolor{red}{red region}) instead of relevant features (\textcolor{blue}{blue region})}. With \rrt, the model can be guided away from these misleading patterns, whether they appear in the \textit{spatial} or \textit{frequency} domain. For this, \rrt leverages feedback on \textcolor{red}{incorrect explanations} to steer the model toward more meaningful and \textcolor{blue}{reliable reasoning}.
    }
    \label{fig:hero}
\end{figure}

Time series data is ubiquitous in today's world. Everything that is measured over time generates some form of time series, for example, energy load \cite{koprinskaConvolutionalNeuralNetworks2018}, sensor measurements in industrial machinery \cite{mehdiyevTimeSeriesClassification2017} or recordings of traffic data \cite{maShortTermTrafficFlow2022}. 
Complex time series data is often analyzed using various neural models \cite{benidisDeepLearningTime2023,ruizGreatMultivariateTime2021}.
However, as in other domains, these can be subject to spurious factors ranging from simple noise or artifacts to complex shortcuts \cite{lapuschkinUnmaskingCleverHans2019a}. Intuitively, a shortcut, also called ``Clever-Hans'' moment, is a spurious pattern in the data that correlates with the target task during training but lacks true relevance. 
If a model learns to rely on such patterns rather than meaningful features, its generalizability suffers, performing well on data with the shortcut but failing on data without it, which poses a significant challenge in real-world deployment \cite{geirhosShortcutLearningDeep2020}. While model explanations can help uncover these shortcuts, they do not resolve the issue on their own (cf. \autoref{fig:hero}~\textbf{I}). Despite extensive research in other domains \cite{steinmannNavigatingShortcutsSpurious2024}, shortcuts in time series models remain underexplored. Existing studies often have specific assumptions about settings and data \cite{bicaTimeSeriesDeconfounder}, leaving a gap in understanding and mitigating shortcut learning in broader time series applications.
To address this, we introduce \rrtfull (\rrt), a new method grounded in the principles of explanatory interactive learning (XIL) \cite{tesoExplanatoryInteractiveMachine2019}, which leverages feedback on explanations to mitigate shortcuts (cf. \autoref{fig:hero} \textbf{II}). \rrt uses traditional explanation methods, such as Integrated Gradients (IG)~\cite{Sundararajan2017AxiomaticAF}, to assess whether the model attends to the correct time steps. It then incorporates feedback on shortcut areas to refine the model, improving robustness and generalization.

However, spurious factors in time series data extend beyond the time domain. For example, a consistent noise frequency in an audio signal can act as a shortcut without being tied to a specific point in time. \rrt can handle these types of shortcuts by incorporating feedback in the frequency domain. To highlight the importance of shortcuts in time series data, we introduce a new real-world dataset with naturally occurring shortcuts, called \textsc{\datafull} (\data). The dataset includes sensor measurements from an industrial high-speed press, essential to many manufacturing processes in the sheet metal working industry. The sensor data for detecting faulty production contains shortcuts and thus provokes incorrect predictions after training. Next to its industrial relevance, \data is the first time series dataset that contains explicitly annotated shortcuts, enabling the evaluation of mitigation strategies on real data.

Altogether, we make the following contributions: (1) We show both on our newly introduced real-world dataset \data and on several other datasets with manual shortcuts that SOTA neural networks on time series classification and forecasting are affected by these shortcuts. (2) We introduce \rrt to mitigate shortcuts for time series data. The method can incorporate feedback on the time domain and the frequency domain. (3) 
By incorporating explanations and feedback in the frequency domain, we enable a new perspective on XIL, overcoming the important limitation that shortcuts must be spatially separable. %

The paper is structured as follows: \autoref{sec:related_work} provides a brief overview of related work on explaining time series and correcting model mistakes. \autoref{sec:method} introduces our approach, while \autoref{sec:shortcuts} describes our decoy methods and \data. We then present a detailed evaluation and discussion in \autoref{sec:exp}. Finally, \autoref{sec:conclusion} concludes the paper and outlines directions for future research.
\section{Related Work}\label{sec:related_work}

\textbf{Explanations for Time Series.}
Explainable artificial intelligence offers various techniques to interpret machine learning models, many of which originated in image or text data before being adapted for time series \cite{rojatExplainableArtificialIntelligence2021}. Attribution methods explain models directly in the input space, while approaches like symbolic aggregation \cite{linSymbolicRepresentationTime2003} and shapelets \cite{yeTimeSeriesShapelets2011} provide higher-level insights (cf. \cite{rojatExplainableArtificialIntelligence2021,schlegelRigorousEvaluationXAI2019} for a broader discussion on time series explanations).
While explanation methods help identify shortcuts, they alone do not enable model revision. Thus, our approach begins with explanations to detect shortcuts and integrates feedback to mitigate them. Specifically, we use Integrated Gradients (IG) \cite{Sundararajan2017AxiomaticAF}, which computes attributions via model gradients and is widely used for time series data \cite{mercierTimeFocusComprehensive2022,veerappaValidationXAIExplanations2022}.

\textbf{Explanatory Interactive Learning (XIL).}
Research on shortcuts and their mitigation is growing, though it primarily focuses on visual data \cite{steinmannNavigatingShortcutsSpurious2024}. One direction is explanatory interactive learning (XIL), which entails methods that revise a model’s decision-making based on human feedback \cite{schramowskiMakingDeepNeural2020,tesoExplanatoryInteractiveMachine2019}. A core aspect of XIL is using model explanations to correct mistakes, particularly to prevent Clever-Hans-like behavior, where models rely on spurious shortcuts \cite{friedrichTypologyExploringMitigation2023,Stammer2020RightFT}.
Several XIL methods have been applied to image data. Right for the Right Reasons (RRR) \cite{Ross2017RightFT} and Right for Better Reasons \cite{shaoRightBetterReasons2021} penalize incorrect attributions, while HINT \cite{selvarajuTakingHINTLeveraging2019} rewards correct focus and \cite{Friedrich2023OneED} explore using multiple explainers. Despite their success in vision tasks, XIL approaches remain largely unexplored for time series. To address this, we introduce \rrt, adapting XIL principles to the unique challenges of time series data.

\textbf{Unconfounding Time Series.}
Apart from interactive learning approaches, some methods address confounding in time series models through causal inference \cite{flandersMethodDetectionResidual2011}. Techniques like the Time Series Deconfounder \cite{bicaTimeSeriesDeconfounder}, SqeDec \cite{hattSequentialDeconfoundingCausal2024}, and LipCDE \cite{caoEstimatingTreatmentEffects2023} estimate data while mitigating confounders in covariates of the target variable. They rely on causal analysis and specific assumptions about data generation.
In contrast, our method focuses on shortcuts within the target variable itself, requiring no assumptions beyond the shortcut being detectable in model explanations - an area where existing causal methods are less applicable.
\section{\rrtfull (\rrt)} \label{sec:method}

\begin{figure}[t]
    \centering    
    \includegraphics[width=0.8\linewidth]{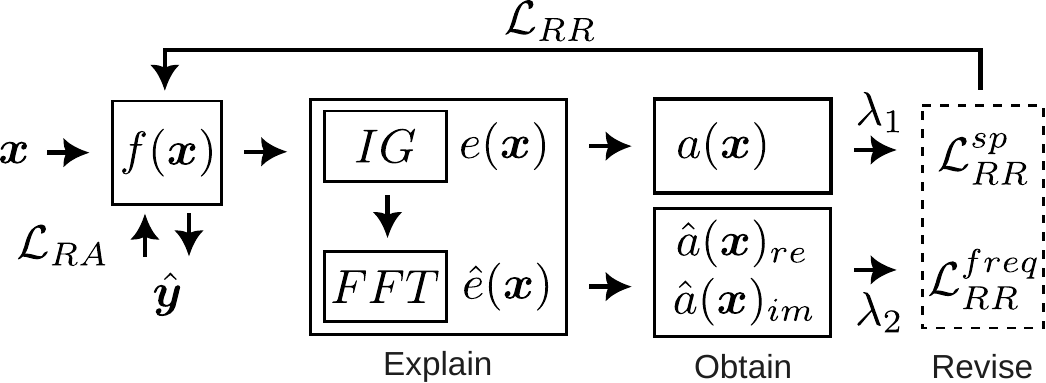}
    \caption{
    \textbf{\methodnames explanation-based revision process.} Input data $\bm{x}$ passes through the model \(f(\bm{x})\) to generate explanations \(e(\bm{x})\), receives annotated feedback \( a(\bm{x}) \), and is then fed back into the model. Integrated Gradients (IG) provides spatial explanations, while Fourier Transform (FFT) converts them into frequency-based explanations. Annotations can be applied in either or both domains and are leveraged by the right-reason loss (\(\mathcal{L}_{RR}^{sp}\) and \(\mathcal{L}_{RR}^{freq}\)) to steer the model away from shortcuts in time or frequency.
    }
    \label{fig:arch_overview}
\end{figure}

The core idea of \rrtfull (\rrt) is to use feedback on model explanations to guide the model away from incorrect reasoning. Following the XIL paradigm, \rrt is designed for seamless integration with other XIL methods. To ensure compatibility, we structure \rrt around the four key steps identified by \cite{friedrichTypologyExploringMitigation2023}: \textit{Select}, \textit{Explain}, \textit{Obtain} and \textit{Revise}.
In \textit{Select}, samples for feedback and model revision are selected. Following previous methods, we select all samples by default but also explore using subsets of the data.
Afterwards, \textit{Explain} covers model explanations before feedback is provided in \textit{Obtain}. Lastly, in \textit{Revise}, the feedback is integrated into the model to overcome the shortcuts.
We introduce \rrt along these steps in the following (as illustrated in \autoref{fig:arch_overview}).

Given a dataset $(\mathcal{X}, \mathcal{Y})$ and a model $f(\cdot)$ for time series classification or forecasting. The dataset consists of $D$ many pairs of $\bm{x}$ and $\bm{y}$. Thereby, $\bm{x} \in \mathcal{X}$ is a time series of length $T$, i.e., $\bm{x}\in\mathbb{R}^T$. For $K$ class classification, the ground-truth output is $\bm{y} \in \{1, \dots, K\}$ and for forecasting, the ground-truth output is the forecasting window $\bm{y} \in \mathbb{R}^W$ of length $W$. The ground-truth output of the full dataset is described as $\mathcal{Y}$ in both cases.
For a datapoint $\bm{x}$, the model generates the output $\hat{\bm{y}} = f(\bm{x})$, where the dimensions of $\hat{\bm{y}}$ are the same as of $\bm{y}$.

\subsection{Explain}
Given a pair of input $\bm{x}$ and model output $\hat{\bm{y}}$ for time series classification, the explainer generates an explanation $e_f(\bm{x}) \in \mathbb{R}^T$ in the form of attributions to explain $\hat{\bm{y}}$ w.r.t.~$\bm{x}$, where a large attribution value means a large influence on the output. In the remainder of the paper, explanations refer to the model $f$, but we drop $f$ from the notation to declutter it, resulting in $e(\bm{x})$.
We use IG~\cite{Sundararajan2017AxiomaticAF} (\autoref{eq:ig}) as an explainer, an established gradient-based attribution method. This method integrates the gradient along the path (using the integration variable $\alpha$) from a baseline $\bar{\bm{x}}$ to the input $x$. It multiplies the result with the difference between baseline and input. However, we make some adjustments to the base method to make it more suitable for time series and model revision, namely taking the absolute value of the difference between $\bm{x}$ and $\bar{\bm{x}}$ (further details in \autoref{sec:app_model_details}). In the following, we introduce the modifications to use attributions for forecasting and to obtain explanations in the frequency domain.

\begin{minipage}[h]{0.6\textwidth}
    \begin{equation}\label{eq:ig}
e(\bm{x})= |\bm{x} - \bar{\bm{x}}| \cdot \int_{0}^{1} \frac{\partial f(\tilde{\bm{x}})}{\partial \tilde{\bm{x}}} \Biggr|_{\tilde{\bm{x}} = \bar{\bm{x}} + \alpha(\bm{x} - \bar{\bm{x}})} d\alpha 
\end{equation}
\end{minipage}
\hfill
\begin{minipage}[h]{0.35\textwidth}
    \begin{equation}\label{eq:ig_forecasting}
    e(\bm{x}) = \frac{1}{W}\sum_{i = 1}^W e'_i(\bm{x})
    \end{equation}
\end{minipage}

\vspace{0.3cm}
\textbf{Attributions for Forecasting.} In a classification setting, attributions are generated by propagating gradients back from the model output (of its highest activated class) to the model inputs. However, there is often no single model output in time series forecasting. Instead, the model simultaneously generates one output for each timestep of the forecasting window. Naively, one could use these $W$ outputs and generate as many explanations $e'_1(\bm{x}), \dots e'_W(\bm{x})$, where each $e'_i(\bm{x})$ is the IG explanation using the i-th time step from the forecasting window as a target instead of a classification label. This number of explanations would, however, make it even harder for humans to interpret the results, as the size of the explanation increases with $W$ \cite{millerExplanationArtificialIntelligence2019}. Therefore, we propose aggregating the individual explanations by averaging in \autoref{eq:ig_forecasting}.
Averaging attributions over the forecasting window provides a simple yet robust aggregation of the explanations. Other means of combining them, potentially even weighted based on distance of the forecast in the future are also imaginable. Overall, this allows attributions for time series classification and forecasting to be generated similarly.

\textbf{Attributions in the Frequency Domain.}
Time series data is often given in the frequency representation, and this format can sometimes be more intuitive for humans to understand than the typical spatial representation. Thus, providing explanations in this domain is essential. \cite{vielhabenExplainableAITime2023} showed how to obtain frequency attributions of the method Layerwise Relevance Propagation \cite{bachPixelWiseExplanationsNonLinear2015}, even if the model does not operate directly on the frequency domain. We adapt this idea to IG: for an input sample $\bm{x}$, we generate attributions with IG, resulting in $e(\bm{x}) \in \mathbb{R}^T$ (\autoref{eq:ig} for classification or \autoref{eq:ig_forecasting} for forecasting). We then interpret the explanation as a time series, with the attribution scores as values. To obtain the frequency explanation, we perform a Fourier transformation of $e(\bm{x})$, resulting in the frequency explanation $\hat{e}(\bm{x}) \in \mathbb{C}^T$ with $\hat{E}$ for the entire set.

\subsection{Obtain}
The next step of \rrt is to obtain feedback on shortcuts. For an input $\bm{x}$, feedback marks input parts via a binary mask $a(\bm{x}) \in \{0, 1\}^T$, where a $1$ signals a potential shortcut at this time step. Thereby, masks $a(\bm{x}) = (0, \dots, 0)^T$ corresponds to no feedback for a sample.
Similarly, feedback can also be given on the frequency explanation, marking which elements in the frequency domain are potential shortcuts. The resulting feedback mask $\hat{a}(\bm{x}) = (\hat{a}(\bm{x})_{re}, \hat{a}(\bm{x})_{im})$ can be different for the real $\hat{a}(\bm{x})_{re} \in \{0, 1\}^T$ and imaginary part $\hat{a}(\bm{x})_{im} \in \{0, 1\}^T$. For the whole dataset, we then have spatial annotations $A$ and frequency annotations $\hat{A}$. 
Obtaining annotated feedback masks can become costly, particularly if the feedback comes from human experts. However, as shortcuts often occur systematically, it can be possible to apply annotations to many samples, drastically reducing the number of annotations required in practice.

\subsection{Revise}
The last step of \rrt is integrating the feedback into the model. We apply the general idea of using a loss-based model revision \cite{Ross2017RightFT,schramowskiMakingDeepNeural2020,Stammer2020RightFT} based on the explanations and the annotation mask. Given the input data $(\mathcal{X}, \mathcal{Y})$, we define the original task (or right-answer) loss as $\mathcal{L}_{RA}(\mathcal{X}, \mathcal{Y})$. This loss measures the model performance and is the primary learning objective. To incorporate the feedback, we further use the right-reason loss $\mathcal{L}_{RR}(A, E)$. This loss aligns model explanations $E = \{e(\bm{x}) | \bm{x} \in \mathcal{X}\}$ and user feedback $A$ by penalizing the model for explanations in the annotated areas. To achieve model revision and a good task performance, both losses are combined, where $\lambda$ is a hyperparameter to balance both parts of the combined loss \mbox{$\mathcal{L}(\mathcal{X}, \mathcal{Y}, A, E) = \mathcal{L}_{\mathrm{RA}}(\mathcal{X}, \mathcal{Y}) + \lambda \mathcal{L}_{\mathrm{RR}}(A, E)$}. Together, the combined loss simultaneously optimizes the primary training objective (e.g. accuracy) and feedback alignment.

\textbf{Time Domain Feedback.}
Masking parts of the time domain is an easy way to mitigate spatially locatable shortcuts (\autoref{fig:hero}, left). We use the explanations $E$ and annotations $A$ in the spatial version of the right-reason loss:
\begin{equation}\label{eq:lrr_spatial}
    \mathcal{L}^{sp}_{RR}(A, E) = \frac{1}{D}\sum_{\bm{x} \in \mathcal{X}} (e(\bm{x}) * a(\bm{x}))^2
\end{equation}
As the explanations and the feedback masks are element-wise multiplied, this loss minimizes the explanation values in marked parts of the input. This effectively trains the model to disregard the marked parts for its computation. Thus, using the loss in \autoref{eq:lrr_spatial} as right-reason component for the combined loss allows to effectively steer the model away from points or intervals in time. 

\textbf{Frequency Domain Feedback.}
However, feedback in the time domain is insufficient to handle every type of shortcut. If the shortcut is not locatable in time, giving spatial feedback cannot be used to revise the models' behavior. Therefore, we utilize explanations and feedback in the frequency domain to tackle shortcuts like in \autoref{fig:hero}, (right). Given the frequency explanations $\hat{E}$ and annotations $\hat{A}$, the right-reason loss for the frequency domain is:
\begin{equation}
    \mathcal{L}^{fr}_{RR}(\hat{A}, \hat{E}) = \frac{1}{D}\sum_{\bm{x} \in \mathcal{X}} \Bigl( (\text{Re}(\hat{e}(\bm{x})) * \hat{a}_{re}(\bm{x}))^2 + (\text{Im}(\hat{e}(\bm{x})) * \hat{a}_{im}(\bm{x}))^2\Bigr)
\end{equation}
The Fourier transformation is invertible and differentiable, so we can backpropagate gradients to parameters directly from this loss. Intuitively, the frequency right-reason loss causes the masked frequency explanations of the model to be small while not affecting any specific point in time.

Depending on the problem, it is possible to use \rrt either in the spatial or frequency domain. Moreover, it is also possible to combine feedback in both domains, including two right-reason terms in the final loss. This results in two parameters $\lambda_1$ and $\lambda_2$ to balance the right-answer and both right-reason losses.
\begin{equation}\label{eq:loss_combined_both}
    \mathcal{L}(\mathcal{X}, \mathcal{Y}, A, E) = \mathcal{L}_{\mathrm{RA}}(\mathcal{X}, \mathcal{Y}) + 
    \lambda_1 \mathcal{L}^{sp}_{\mathrm{RR}}(A, E) + \lambda_2 \mathcal{L}^{fr}_{\mathrm{RR}}(\hat{A}, \hat{E})
\end{equation}
It is important to note that giving feedback in the frequency domain allows a new form of model revision through XIL. Even if we effectively still apply masking in the frequency domain, the effect in the original input domain is entirely different. Masking out a single frequency affects all time points without preventing the model from looking at any of them. In general, having an invertible transformation from the input domain to a different representation allows to give feedback more flexible than before. The Fourier transformation is a prominent example but not the only possible choice for this. Using other transformations like wavelets \cite{grapsIntroductionWavelets1995}, is also possible. 

\textbf{Computational Costs.} Including \rrt in the training of a model increases the computational cost. Computing the right reason loss term requires the computation of a mixed partial derivative: $\frac{\partial^2 f_\theta(x)}{\partial \theta \partial x}$. Even though this is a second-order derivative, it does not result in any substantial cost increases, as the second-order component of our loss can be formalized as a Hessian-vector product (cf.~\autoref{sec:app_cost}), which is known to be fast to compute \cite{martensDeepLearningHessianfree2010}. We also observed this in our experimental evaluation, as even the naive implementation of our loss in PyTorch scales to large models.

\textbf{Source of Feedback.}
A key aspect of \rrt is the feedback incorporated in the \textit{Obtain} step, which can come from various sources, including automated methods, rule-based systems, foundation models, or human annotations. Automated approaches, such as rule-based heuristics or pre-trained foundation models, provide scalable and consistent feedback, reducing the reliance on manual labeling. However, human annotations remain valuable for ensuring accuracy, especially in complex cases where automated methods may introduce biases or fail to capture nuanced patterns. \rrt is agnostic to the feedback source, allowing flexibility in its application. 

\section{Shortcuts in Time Series}\label{sec:shortcuts}
Shortcuts, like those in images, naturally occur in time series data but are often less apparent. Developing effective mitigation methods requires datasets where shortcuts are explicitly annotated, yet no existing datasets provide such annotations, despite known biases in popular datasets \cite{bagnallTransformationBasedEnsembles2012}. 
To address this gap, we introduce several time series dataset decoy variants inspired by prior work on decoy data \cite{Ross2017RightFT}, allowing for controlled evaluation of shortcut mitigation strategies. To further evaluate shortcut mitigation under real-world conditions, we present \data, a real-world dataset where shortcuts arise from sensor recording processes. 

\subsection{Decoy Shortcuts}
\textbf{Classification.}
For both spatial and frequency cases, we introduce the shortcut as a class-specific pattern embedded in each training sample. The \textbf{spatial} pattern \textit{replaces} \(m\) time steps with a sine wave defined as \(s := \sin(t \cdot (2 + j)\pi) \cdot A\)
where \( t \in \{0, 1, \dots, m\} \) are the respective time steps, \( j \) represents the class index and \(A\) is the amplitude. In contrast, the \textbf{frequency} pattern is a similar sine wave, but it is \textit{additively} applied to the full sequence (\(m = T\)).

\textbf{Forecasting.}
For forecasting datasets, spatial decoys are more challenging due to window-based sampling and the complexity of the target. To address this, we design the shortcut as a "back-copy" of the forecast, where the decoy is equivalent to the actual solution. Due to the windowed sampling, the shortcut appears in every second sample. Given a sample of \emph{lookback window} \(\bm{x}\in \mathbb{R}^T\) and the \emph{forecast horizon} \(\bm{y}\in \mathbb{R}^W\). In the shortcut samples, we overwrite the first \(W\) entries of the lookback window with the future horizon values, yielding:
\[
\bm{x}^{\mathrm{s}}
=
\Bigl(
\underbrace{y_1,\, \dots,\, y_W}_{\text{future horizon}},\quad
\underbrace{x_{W+1},\, \dots,\, x_{T}}_{\text{remaining lookback}}
\Bigr).
\]
while \(y\) remains unchanged. 
While this setting may seem constructed, similar patterns can emerge in real-world scenarios. For instance, in data transmission, glitches such as packet losses or duplications can subtly introduce irregularities into time series data, inadvertently creating forecasting shortcuts.

We model the forecasting frequency decoy as a recurring Dirac impulse with a specific frequency, added every $k$ time steps: 
$ i \in \{n \cdot k | n \in \mathbb{N} \wedge n \cdot k \leq T+W\} $ with a strength of $A$:
\( \mathit{interference} := A \cdot \Delta_{i} \). The impulse is present within the lookback and forecasting window during training, representing an effective decoy distracting the model from the actual forecast.

\subsection{Real-World Shortcuts: \datafull (\data)}
\rrt aims to mitigate shortcuts in time series data. While the decoys above provide a controlled evaluation setting, they do not capture the complexity of real-world shortcuts. To rectify this, we introduce \textsc{\datafull (\data)}\footnote{\url{https://huggingface.co/datasets/AIML-TUDA/P2S}}, a dataset of sensor recordings with naturally occurring shortcuts. 

\begin{figure}[t]
\centering
    \includegraphics[width=1\linewidth]{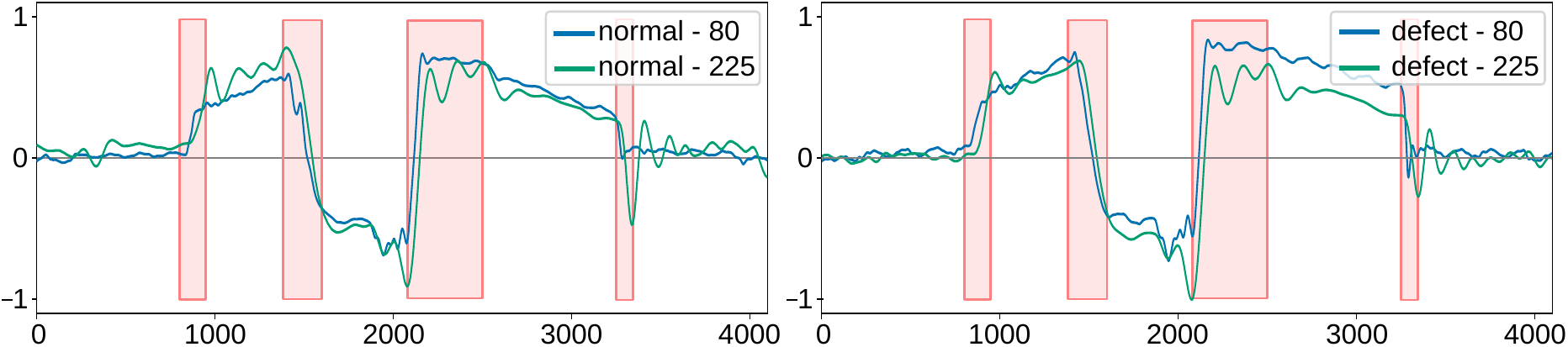}
    \caption{\textbf{Samples of \data from normal (left) and defect (right) class at 80 and 225 strokes per minute.} Areas of the time series that are especially sensitive to the stroke rate are considered a shortcut and marked \textcolor{red}{red}.}
\label{fig:p2s_example}
\end{figure}

The sensor data stems from a high-speed press production line for metal parts, one of the sheet metal working industry's most economically significant processes. Based on the sensor data, the task is to predict whether a run is defective. The recordings include different production speeds, which, although not affecting part quality, influence process friction, and applied forces. \autoref{fig:p2s_example} shows samples recorded at different speeds from normal and defect runs, highlighting variations even within the same class. 
A domain expert identified regions in the time series that vary with production speed, potentially distracting models from relevant features, especially when no defect and normal runs of the same speed are in the training data. In these cases, the run’s speed is a shortcut and makes it difficult to generalize to other speeds than those present in training. \data also includes a specifically curated setup that matches run speeds during training to avoid the shortcut. Further details on the dataset are available in \autoref{sec:app_dataset}.

\section{Experimental Evaluations} \label{sec:exp}
In this section, we investigate the effectiveness of \rrt\codenote to mitigate shortcuts in time series classification and forecasting, including revision in the spatial domain (\rrts) and the frequency domain (\rrtf), as well as both jointly.

\subsection{Experimental Setup}

\textbf{Data.} 
For classification, we use datasets from the UCR/UEA repository~\cite{ucr_archive18}. We select available datasets of a minimal size (cf.~\autoref{sec:app_ucr}), which results in \textsc{Fault Detection A}, \textsc{Ford A}, \textsc{Ford B}, and \textsc{Sleep}. 
For time series forecasting, we evaluate on three popular datasets from the Darts repository \cite{herzenDartsUserFriendlyModern2022}: \textsc{ETTM1}, \textsc{Energy}, and \textsc{Weather}.
We split the data into training and test sets using a 70/30 ratio and 20\% of the training set are used for validation.
We apply the previously described decoys to the training sets and simulate feedback based on the shortcuts to generate annotation masks. In the real-world experiment, we utilize our newly introduced dataset \data .
The mask is applied to all samples except in our feedback scaling experiment. For the real-world test case, we consider our newly introduced dataset \data. We standardize all datasets as suggested by \cite{wu2021autoformer}, i.e., rescaling the distribution to zero mean and a standard deviation of one.

\textbf{Models.}
For time series classification, we use the FCN model of \cite{ma2023survey}, with a slightly modified architecture for Sleep to achieve better performance (cf. ~\autoref{sec:app_model_details}). Additionally, we use the OFA model  \cite{zhou2023one}.
For forecasting, we use TiDE  \cite{dasLongtermForecastingTiDE2023}, PatchTST \cite{Yuqietal-2023-PatchTST} and NBEATS \cite{oreshkin_NBEATS} to highlight the applicability of our method to a variety of model classes.

\textbf{Metrics.}
In our evaluations, we compare model performance on datasets with and without shortcuts, as well as with and without \rrt. For classification, we report balanced (multiclass) accuracy (ACC), and mean squared error (MSE) for forecasting. The respective mean absolute error (MAE) results can be found in~\autoref{sec:app_add_res}. We report average and standard deviation over 5 runs.
\begin{table*}[t]
\centering
\scriptsize
\caption{\textbf{Applying RioT mitigates shortcuts in time series classification.} Performance before and after applying \rrt for spatial (Base\textsubscript{sp}) and frequency (Base\textsubscript{freq}) shortcuts. High training and low test accuracies indicate overfitting to the shortcut, which \rrt successfully mitigates. \textit{No Shortcut} represents the ideal scenario where the model is not affected by any shortcut.
}
\label{tab:xil_tsc_full}
\setlength{\tabcolsep}{5.1pt}
\vskip 0.1in
\resizebox{\columnwidth}{!}{%
\begin{tabular}{lll@{\hspace{1pt}}cl@{\hspace{1pt}}cl@{\hspace{1pt}}cl@{\hspace{1pt}}c}
\toprule
Model & Config  & \multicolumn{2}{c}{Fault Detection A} & \multicolumn{2}{c}{FordA} & \multicolumn{2}{c}{FordB} & \multicolumn{2}{c}{Sleep} \\
 &  (ACC $\uparrow$) & Train & \makebox[5pt][r]{Test} & Train & \makebox[5pt][r]{Test} & Train & \makebox[5pt][r]{Test} & Train & \makebox[5pt][r]{Test} \\
\midrule
FCN & \tiny No Shortcut & 0.99 \textpm \tiny 0.00 & \multicolumn{1}{r}{0.99 \textpm \tiny 0.00} & 0.92 \textpm \tiny 0.01 & \multicolumn{1}{r}{0.91 \textpm \tiny 0.00} & 0.93 \textpm \tiny 0.00 & \multicolumn{1}{r}{0.76 \textpm \tiny 0.01} & 0.68 \textpm \tiny 0.00 & \multicolumn{1}{r}{0.62 \textpm \tiny 0.00} \\
\cmidrule{2-10}\morecmidrules\cmidrule{2-10}
 & Base\textsubscript{sp} & \textbf{1.00} \textpm \tiny 0.00 & \multicolumn{1}{r}{0.74 \textpm \tiny 0.06} & \textbf{1.00} \textpm \tiny 0.00 & \multicolumn{1}{r}{0.71 \textpm \tiny 0.08} & \textbf{1.00} \textpm \tiny 0.00 & \multicolumn{1}{r}{0.63 \textpm \tiny 0.03} & \textbf{1.00} \textpm \tiny 0.00 & \multicolumn{1}{r}{0.10 \textpm \tiny 0.03} \\
 & + RioT\textsubscript{sp} & 0.98 \textpm \tiny 0.01 & \multicolumn{1}{r}{\textbf{0.93} \textpm \tiny 0.03} & 0.99 \textpm \tiny 0.01 & \multicolumn{1}{r}{\textbf{0.84} \textpm \tiny 0.02} & 0.99 \textpm \tiny 0.00 & \multicolumn{1}{r}{\textbf{0.68} \textpm \tiny 0.02} & 0.60 \textpm \tiny 0.06 & \multicolumn{1}{r}{\textbf{0.54} \textpm \tiny 0.05} \\
\cmidrule{2-10}
 & Base\textsubscript{freq} & \textbf{0.98} \textpm \tiny 0.01 & \multicolumn{1}{r}{0.87 \textpm \tiny 0.03} & \textbf{0.98} \textpm \tiny 0.00 & \multicolumn{1}{r}{0.73 \textpm \tiny 0.01} & \textbf{0.99} \textpm \tiny 0.01 & \multicolumn{1}{r}{0.60 \textpm \tiny 0.01} & \textbf{0.98} \textpm \tiny 0.00 & \multicolumn{1}{r}{0.27 \textpm \tiny 0.02} \\
 & + RioT\textsubscript{freq} & 0.94 \textpm \tiny 0.00 & \multicolumn{1}{r}{\textbf{0.90} \textpm \tiny 0.03} & 0.83 \textpm \tiny 0.02 & \multicolumn{1}{r}{\textbf{0.83} \textpm \tiny 0.02} & 0.94 \textpm \tiny 0.00 & \multicolumn{1}{r}{\textbf{0.65} \textpm \tiny 0.01} & 0.67 \textpm \tiny 0.05 & \multicolumn{1}{r}{\textbf{0.45} \textpm \tiny 0.07} \\
\midrule
OFA & \tiny No Shortcut & 1.00 \textpm \tiny 0.00 & \multicolumn{1}{r}{0.98 \textpm \tiny 0.02} & 0.92 \textpm \tiny 0.01 & \multicolumn{1}{r}{0.87 \textpm \tiny 0.04} & 0.95 \textpm \tiny 0.01 & \multicolumn{1}{r}{0.70 \textpm \tiny 0.04} & 0.69 \textpm \tiny 0.00 & \multicolumn{1}{r}{0.64 \textpm \tiny 0.01} \\
\cmidrule{2-10}\morecmidrules\cmidrule{2-10}
 & Base\textsubscript{sp} & \textbf{1.00} \textpm \tiny 0.00 & \multicolumn{1}{r}{0.53 \textpm \tiny 0.02} & \textbf{1.00} \textpm \tiny 0.00 & \multicolumn{1}{r}{0.50 \textpm \tiny 0.00} & \textbf{1.00} \textpm \tiny 0.00 & \multicolumn{1}{r}{0.52 \textpm \tiny 0.01} & \textbf{1.00} \textpm \tiny 0.00 & \multicolumn{1}{r}{0.21 \textpm \tiny 0.05} \\
 & + RioT\textsubscript{sp} & 0.96 \textpm \tiny 0.08 & \multicolumn{1}{r}{\textbf{0.98} \textpm \tiny 0.01} & 0.92 \textpm \tiny 0.03 & \multicolumn{1}{r}{\textbf{0.85} \textpm \tiny 0.02} & 0.94 \textpm \tiny 0.01 & \multicolumn{1}{r}{\textbf{0.65} \textpm \tiny 0.04} & 0.52 \textpm \tiny 0.22 & \multicolumn{1}{r}{\textbf{0.58} \textpm \tiny 0.05} \\
\cmidrule{2-10}
 & Base\textsubscript{freq} & \textbf{1.00} \textpm \tiny 0.00 & \multicolumn{1}{r}{0.72 \textpm \tiny 0.02} & \textbf{1.00} \textpm \tiny 0.00 & \multicolumn{1}{r}{0.65 \textpm \tiny 0.01} & 1.00 \textpm \tiny 0.00 & \multicolumn{1}{r}{0.56 \textpm \tiny 0.02} & \textbf{0.99} \textpm \tiny 0.00 & \multicolumn{1}{r}{0.24 \textpm \tiny 0.03} \\
 & + RioT\textsubscript{freq} & 0.96 \textpm \tiny 0.02 & \multicolumn{1}{r}{\textbf{0.98} \textpm \tiny 0.02} & 0.78 \textpm \tiny 0.04 & \multicolumn{1}{r}{\textbf{0.85} \textpm \tiny 0.04} & \textbf{1.00} \textpm \tiny 0.00 & \multicolumn{1}{r}{\textbf{0.64} \textpm \tiny 0.03} & 0.50 \textpm \tiny 0.16 & \multicolumn{1}{r}{\textbf{0.49} \textpm \tiny 0.04} \\
\bottomrule
\end{tabular}
}

\end{table*}

\subsection{Evaluations}

\textbf{Removing Shortcuts for Time Series Classification.}
We evaluate the effectiveness of \rrt (spatial: \rrts, frequency: \rrtf) in addressing shortcuts in classification tasks by comparing balanced accuracy with and without \rrt.
As shown in \autoref{tab:xil_tsc_full}, without \rrt, both FCN and OFA overfit to shortcuts, achieving $\approx$100\% training accuracy, while having poor test performance. Applying \rrt significantly improves test performance for both models across all datasets. In some cases, RioT even reaches the performance of the ideal reference (no shortcut) scenario as if there would be no shortcut in the data. Even on FordB, where the drop in training-to-test performance highlights the distribution shift of that dataset \cite{bagnallTransformationBasedEnsembles2012}, \rrts is still beneficial. 
Similarly, \rrtf enhances performance on data with frequency shortcuts, though the improvement is less pronounced for FCN on Ford B, suggesting essential frequency information is sometimes obscured by \rrtf.
In summary, \rrt successfully mitigates shortcuts in both domains, enhancing test generalization for FCN and OFA models.

\textbf{Removing Shortcuts for Time Series Forecasting.}
Shortcuts are not exclusive to time series classification and can significantly impact other tasks, such as forecasting. In \autoref{tab:xil_tsf_full}, we outline that spatial shortcuts cause models to overfit, but applying \rrts reduces MSE across datasets, especially for Energy, where MSE drops by up to 56\%. 
In the frequency-shortcut setting, the training data includes a recurring Dirac impulse as a decoy (cf.~\autoref{sec:conf_details} for details). \rrtf alleviates this distraction and improves the test performance significantly. For example, TiDE's test MSE on ETTM1 decreases by 14\% compared to the decoy setting.

In general, \rrt effectively addresses spatial and frequency shortcuts in forecasting tasks. Interestingly, for TiDE on the Energy dataset, the performance with \rrtf even surpasses the no shortcut model. Here, the added frequency acts as a form of data augmentation, enhancing model robustness. A similar behavior can be observed for NBEATS and ETTM1, where the decoy setting actually improves the model slightly, and \rrt even improves upon that.
\begin{table*}[t]
\centering
\scriptsize
\caption{\textbf{\rrt can successfully overcome shortcuts in time series forecasting.}
MSE values (MAE values cf. \autoref{tab:xil_tsf_full_mae}) on the training set with and test set without shortcuts. \textit{No Shortcut} is the ideal scenario where the model is not affected by shortcuts.
}
\label{tab:xil_tsf_full}
\setlength{\tabcolsep}{5.1pt}
\vskip 0.1in
\resizebox{\columnwidth}{!}{%
\begin{tabular}{lll@{\hspace{1pt}}cl@{\hspace{1pt}}cl@{\hspace{1pt}}c}
    \toprule
    Model & Config (MSE $\downarrow$) & \multicolumn{2}{c}{ETTM1} & \multicolumn{2}{c}{Energy} & \multicolumn{2}{c}{Weather} \\
     &  & Train & \makebox[5pt][r]{Test} & Train & \makebox[5pt][r]{Test} & Train & \makebox[5pt][r]{Test} \\
    \midrule
    NBEATS & No Shortcut & 0.30 \textpm \tiny 0.02 & \multicolumn{1}{r}{0.47 \textpm \tiny 0.02} & 0.34 \textpm \tiny 0.03 & \multicolumn{1}{r}{0.26 \textpm \tiny 0.02} & 0.08 \textpm \tiny 0.01 & \multicolumn{1}{r}{0.03 \textpm \tiny 0.01} \\
    \cmidrule{2-8}\morecmidrules\cmidrule{2-8}
     & Base\textsubscript{sp} & \textbf{0.24} \textpm \tiny 0.01 & \multicolumn{1}{r}{0.55 \textpm \tiny 0.01} & \textbf{0.33} \textpm \tiny 0.03 & \multicolumn{1}{r}{0.94 \textpm \tiny 0.02} & \textbf{0.09} \textpm \tiny 0.01 & \multicolumn{1}{r}{0.16 \textpm \tiny 0.04} \\
     & + RioT\textsubscript{sp} & 0.30 \textpm \tiny 0.01 & \multicolumn{1}{r}{\textbf{0.50} \textpm \tiny 0.01} & 0.45 \textpm \tiny 0.03 & \multicolumn{1}{r}{\textbf{0.58} \textpm \tiny 0.01} & 0.11 \textpm \tiny 0.01 & \multicolumn{1}{r}{\textbf{0.09} \textpm \tiny 0.02} \\
    \cmidrule{2-8}
     & Base\textsubscript{freq} & \textbf{0.30} \textpm \tiny 0.02 & \multicolumn{1}{r}{0.46 \textpm \tiny 0.01} & \textbf{0.33} \textpm \tiny 0.04 & \multicolumn{1}{r}{0.36 \textpm \tiny 0.04} & \textbf{0.11} \textpm \tiny 0.02 & \multicolumn{1}{r}{0.32 \textpm \tiny 0.09} \\
     & + RioT\textsubscript{freq} & 0.31 \textpm \tiny 0.02 & \multicolumn{1}{r}{\textbf{0.45} \textpm \tiny 0.01} & \textbf{0.33} \textpm \tiny 0.04 & \multicolumn{1}{r}{\textbf{0.34} \textpm \tiny 0.04} & 0.81 \textpm \tiny 0.48 & \multicolumn{1}{r}{\textbf{0.17} \textpm \tiny 0.01} \\
    \midrule
    PatchTST & No Shortcut & 0.46 \textpm \tiny 0.03 & \multicolumn{1}{r}{0.47 \textpm \tiny 0.01} & 0.26 \textpm \tiny 0.01 & \multicolumn{1}{r}{0.23 \textpm \tiny 0.00} & 0.26 \textpm \tiny 0.03 & \multicolumn{1}{r}{0.08 \textpm \tiny 0.01} \\
    \cmidrule{2-8}\morecmidrules\cmidrule{2-8}
     & Base\textsubscript{sp} & \textbf{0.40} \textpm \tiny 0.02 & \multicolumn{1}{r}{0.55 \textpm \tiny 0.01} & \textbf{0.29} \textpm \tiny 0.01 & \multicolumn{1}{r}{0.96 \textpm \tiny 0.03} & \textbf{0.20} \textpm \tiny 0.03 & \multicolumn{1}{r}{0.19 \textpm \tiny 0.01} \\
     & + RioT\textsubscript{sp} & \textbf{0.40} \textpm \tiny 0.03 & \multicolumn{1}{r}{\textbf{0.53} \textpm \tiny 0.01} & 0.44 \textpm \tiny 0.00 & \multicolumn{1}{r}{\textbf{0.45} \textpm \tiny 0.01} & 0.55 \textpm \tiny 0.20 & \multicolumn{1}{r}{\textbf{0.14} \textpm \tiny 0.01} \\
    \cmidrule{2-8}
     & Base\textsubscript{freq} & \textbf{0.45} \textpm \tiny 0.03 & \multicolumn{1}{r}{0.91 \textpm \tiny 0.16} & \textbf{0.04} \textpm \tiny 0.00 & \multicolumn{1}{r}{0.53 \textpm \tiny 0.05} & \textbf{0.63} \textpm \tiny 0.09 & \multicolumn{1}{r}{0.24 \textpm \tiny 0.04} \\
     & + RioT\textsubscript{freq} & 0.91 \textpm \tiny 0.07 & \multicolumn{1}{r}{\textbf{0.66} \textpm \tiny 0.04} & 0.71 \textpm \tiny 0.10 & \multicolumn{1}{r}{\textbf{0.38} \textpm \tiny 0.06} & 0.96 \textpm \tiny 0.02 & \multicolumn{1}{r}{\textbf{0.17} \textpm \tiny 0.00} \\
    \midrule
    TiDE & No Shortcut & 0.27 \textpm \tiny 0.01 & \multicolumn{1}{r}{0.47 \textpm \tiny 0.01} & 0.27 \textpm \tiny 0.01 & \multicolumn{1}{r}{0.26 \textpm \tiny 0.02} & 0.25 \textpm \tiny 0.02 & \multicolumn{1}{r}{0.03 \textpm \tiny 0.00} \\
    \cmidrule{2-8}\morecmidrules\cmidrule{2-8}
     & Base\textsubscript{sp} & \textbf{0.22} \textpm \tiny 0.01 & \multicolumn{1}{r}{0.54 \textpm \tiny 0.03} & \textbf{0.28} \textpm \tiny 0.01 & \multicolumn{1}{r}{1.19 \textpm \tiny 0.03} & \textbf{0.22} \textpm \tiny 0.03 & \multicolumn{1}{r}{0.15 \textpm \tiny 0.01} \\
     & + RioT\textsubscript{sp} & 0.23 \textpm \tiny 0.01 & \multicolumn{1}{r}{\textbf{0.48} \textpm \tiny 0.01} & 0.53 \textpm \tiny 0.02 & \multicolumn{1}{r}{\textbf{0.52} \textpm \tiny 0.02} & 0.25 \textpm \tiny 0.03 & \multicolumn{1}{r}{\textbf{0.11} \textpm \tiny 0.01} \\
    \cmidrule{2-8}
     & Base\textsubscript{freq} & \textbf{0.06} \textpm \tiny 0.01 & \multicolumn{1}{r}{0.69 \textpm \tiny 0.08} & \textbf{0.07} \textpm \tiny 0.01 & \multicolumn{1}{r}{0.34 \textpm \tiny 0.08} & \textbf{0.79} \textpm \tiny 0.09 & \multicolumn{1}{r}{0.31 \textpm \tiny 0.09} \\
     & + RioT\textsubscript{freq} & 0.07 \textpm \tiny 0.01 & \multicolumn{1}{r}{\textbf{0.49} \textpm \tiny 0.07} & \textbf{0.07} \textpm \tiny 0.01 & \multicolumn{1}{r}{\textbf{0.21} \textpm \tiny 0.02} & 1.12 \textpm \tiny 0.36 & \multicolumn{1}{r}{\textbf{0.22} \textpm \tiny 0.01} \\
    \bottomrule
    \end{tabular}
}

\end{table*}

\begin{figure}[t]
    \centering
    \includegraphics[width=\linewidth]{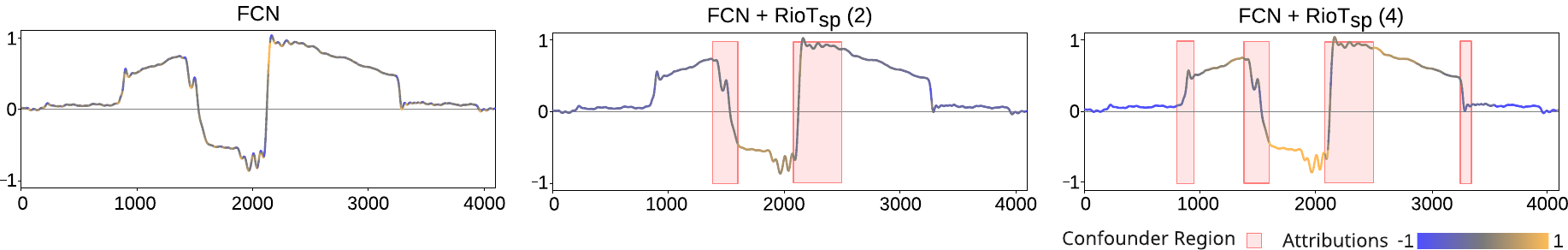}
    \caption{\textbf{Applying \rrt lets the model ignore shortcut areas.} While FCN primarily focuses on shortcuts, applying \rrt with partial feedback (middle) or full feedback (bottom) causes the model to ignore the shortcut and focus on the remaining input.}
    \label{fig:spatial_mechanical}
\end{figure}

\textbf{Removing Shortcuts in the Real-World.}
\label{sec:removing_conf_rw}
So far, our experiments have demonstrated \rrt's ability to counteract shortcuts within controlled environments. However, real-world shortcuts, as in our new dataset \data, often have more complex structures. The model explanations in \autoref{fig:spatial_mechanical} (top) reveal a focus on distinct regions of the sensor curve, specifically the two middle regions. With domain knowledge, it is clear that these regions should not affect the model's output. By applying \rrt, we can redirect the model's attention away from these regions. New model explanations highlight that the model still focuses on (other) incorrect regions, which can be mitigated by extending the annotated area. In 
\autoref{tab:mechanical_data}, the model performance (exemplarly with FCN) in these settings is presented. Without \rrt, the model overfits to the shortcut. the test performance improves already with partial feedback (2) and even more with full feedback (4). These results highlight the effectiveness of \rrt in real-world scenarios where not all shortcuts are initially known.

\textbf{Removing Multiple Shortcuts at Once.}
In the previous experiments, we illustrated that \rrt is suitable for addressing individual shortcuts, whether spatial or frequency-based. However, real-world time series data often presents a blend of multiple shortcuts that simultaneously influence model performance.
\begin{table}[t]
    \begin{minipage}[t]{.43\linewidth}
        \centering
        \setlength{\tabcolsep}{4.1pt}
        \tiny{
      \caption{\textbf{Applying \rrt overcomes shortcuts in \data.} Performance on the train set with and test set without shortcuts. FCN learns the train shortcut, resulting in lower test performance. Applying \rrt with partial feedback (2) already yields good improvements, while adding feedback on the full shortcut area (4) is even better. \textit{No Shortcut} is the ideal scenario, specifically curated so that there is no shortcut.
             }
            \label{tab:mechanical_data}
            \begin{tabular}{lll}
\toprule
P2S (ACC $\uparrow$) & Train & \text{\ \ \ Test} \\
\midrule
FCN\textsubscript{No Shortcut} & 0.97 \textpm \tiny 0.01 & \multicolumn{1}{r}{0.95 \textpm \tiny 0.01} \\
\midrule
\midrule
FCN\textsubscript{sp} & \textbf{0.99} \textpm \tiny 0.01 & \multicolumn{1}{r}{0.66 \textpm \tiny 0.14} \\
FCN\textsubscript{sp} + \rrts (2) & 0.96 \textpm \tiny 0.01 & \multicolumn{1}{r}{0.78 \textpm \tiny 0.05} \\
FCN\textsubscript{sp} + \rrts (4) & 0.95 \textpm \tiny 0.01 & \textbf{0.82} \textpm \tiny 0.06 \\

\bottomrule
\end{tabular}

        }
    \end{minipage}%
    \hfill
    \begin{minipage}[t]{.55\linewidth}
        \setlength{\tabcolsep}{4.1pt}
      \centering
        \tiny{
        \caption{\textbf{\rrt can combine spatial and frequency feedback.}
        If the data contains time and frequency shortcuts, \rrt can combine feedback on both domains to mitigate them, superior to feedback on only one domain. \textit{No Shortcut} represents the ideal scenario when the model is not affected by any shortcuts.
        }
    \label{tab:combined_xil}
    \begin{tabular}{lll}
        \toprule
        Sleep (Classification ACC $\uparrow$)& Train & \text{\ \ \ Test} \\ \midrule
FCN\textsubscript{No Shortcut} & 0.68 \textpm \tiny{0.00} & \multicolumn{1}{r}{0.62 \textpm \tiny{0.00}} \\
        \midrule 
        \midrule 
    FCN\textsubscript{freq,sp} & \textbf{1.00} \textpm \tiny{0.00} & \multicolumn{1}{r}{0.10 \textpm \tiny{0.04}}  \\
FCN\textsubscript{freq,sp} + \rrts & 0.94 \textpm \tiny{0.00}  & \multicolumn{1}{r}{0.24 \textpm \tiny{0.02}} \\
FCN\textsubscript{freq,sp} + \rrtf & \textbf{1.00} \textpm \tiny{0.00} & \multicolumn{1}{r}{0.04 \textpm \tiny{0.00}}  \\
FCN\textsubscript{freq,sp} + \rrtfs & 0.47 \textpm \tiny{0.00}  & \textbf{0.48} \textpm \tiny{0.03}  \\
        \midrule\midrule
        Energy (Forecasting MSE $\downarrow$) & Train & \text{\ \ \ Test} \\ \midrule
TiDE\textsubscript{No Shortcut} & 0.28 \textpm \tiny{0.01}  & \multicolumn{1}{r}{0.26 \textpm \tiny{0.02}}  \\
\midrule \midrule
TiDE\textsubscript{freq,sp} & \textbf{0.16} \textpm \tiny{0.01} & \multicolumn{1}{r}{0.74 \textpm \tiny{0.02}}  \\
TiDE\textsubscript{freq,sp} + \rrts & 0.20 \textpm \tiny{0.01}  & \multicolumn{1}{r}{0.61 \textpm \tiny{0.02}}  \\
TiDE\textsubscript{freq,sp} + \rrtf & 0.22 \textpm \tiny{0.01}  & \multicolumn{1}{r}{0.55 \textpm \tiny{0.02}}  \\
TiDE\textsubscript{freq,sp} + \rrtfs & 0.25 \textpm \tiny{0.01}  & \textbf{0.47} \textpm \tiny{0.01}  \\
\bottomrule
    \end{tabular}

    }
    \end{minipage} 
\end{table}
Thus, we investigate the impact of applying \rrt to both spatial and frequency shortcuts simultaneously (cf.~\autoref{tab:combined_xil}), exemplarily using FCN and TiDE. When Sleep contains shortcuts in both domains, FCN without \rrt overfits and fails to generalize. Addressing only one shortcut does not mitigate the effects, as the model adapts to the other. However, combining the respective feedback from both domains (\rrtfs) significantly improves test performance, matching the frequency-shortcut scenario (cf.~\autoref{tab:xil_tsc_full}). 
\autoref{tab:combined_xil} (bottom) shows the impact of multiple shortcuts on the Energy dataset, where the lower training MSE indicates overfitting. While applying either spatial or frequency feedback individually shows some effect, utilizing both types of feedback simultaneously (\rrtfs) results in the largest improvements, as both decoys are addressed.
The performance gap between \rrtfs and the no shortcut setting is more pronounced than in single shortcut cases (cf. \autoref{tab:xil_tsf_full}). This highlights again the known challenge of removing multiple shortcuts at once, which is generally more complex than individual shortcuts \cite{steinmannNavigatingShortcutsSpurious2024}. 

\begin{figure}[t]
\centering
\begin{minipage}{.48\textwidth}
    \centering
    \includegraphics[width=0.9\linewidth]{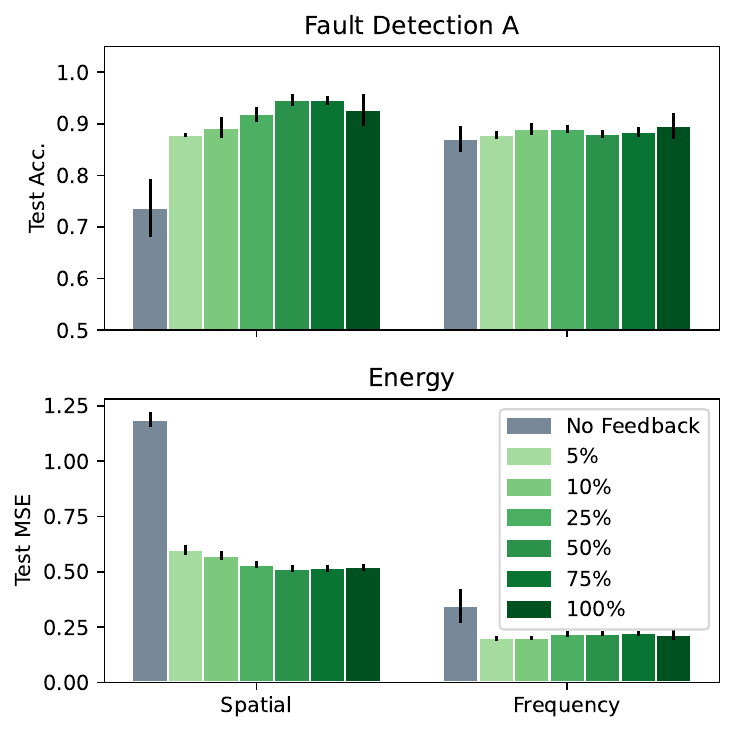}
    \caption{\textbf{RioT uses feedback efficiently.} Even with feedback on only a small percentage of the data, RioT can overcome shortcuts.}
    \label{fig:scaling_exp}
\end{minipage}%
\hfill
\begin{minipage}{.48\textwidth}
    \centering
    \includegraphics[width=0.9\linewidth]{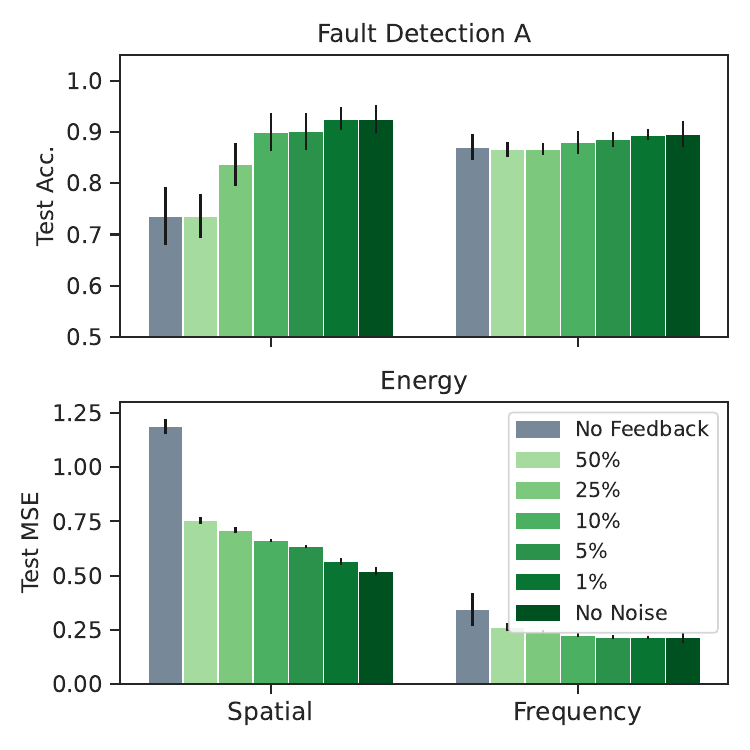}
  \caption{\textbf{\rrt is robust against invalid feedback.} Even with some percentage of random feedback, \rrt overcomes the shortcuts.}
  \label{fig:noisy_exp}
\end{minipage}
\end{figure}

\textbf{Handling Feedback.}
As the annotations are a crucial component of \rrt, we conduct two different ablation studies to evaluate its impact using the classification data set Fault Detection A and the forecasting data set Energy. The first experiment examines the required amount of feedback, while the second assesses robustness to noisy feedback. In particular if the feedback stems from domain experts, making excessive feedback requests is impractical. Thus our first experiment evaluates the performance of \rrt when feedback is provided on only a portion of the dataset (\autoref{fig:scaling_exp}). The findings reveal that full annotations are unnecessary. Even with minimal feedback, such as annotating just 5\% of the samples, \rrt significantly outperforms scenarios with no feedback.
While previous experiments assumed entirely accurate feedback, real-world applications often involve some degree of error. Therefore, we test the resilience of \rrt to increasing levels of incorrect feedback (\autoref{fig:noisy_exp}). Instead of accurately marking shortcut areas, random time steps or frequency components are incorrectly labeled as shortcuts. The results show that \rrt maintains strong performance even with up to 10\% invalid feedback, presenting only slight performance declines. In certain cases, like forecasting with spatial shortcuts, \rrt can still achieve notable improvements despite high levels of feedback noise.
To further evaluate whether annotations in different settings can also be incorporated via \rrt, we conduct an additional ablation where the feedback is based on shaplets instead of the input domain directly
(cf. \autoref{tab:shaplet_results} in the appendix). The results show that \rrt can be effective in this setup as well, and is not limited to the specific explanation method and annotation modality shown in the other experiments (more details in \autoref{sec:app_add_res}).
In summary, \rrt effectively generalizes from small subsets of feedback and remains robust against a moderate amount of annotation noise. Additionally, \rrt can incorporate feedback in other settings with other types of explanations as well. These qualities demonstrate that \rrt is well-equipped to manage the practical challenges associated with incorporating feedback.

\textbf{Qualitative Insights into Model Encodings.}
\begin{figure}[h]
    \centering
    \includegraphics[width=1\linewidth]{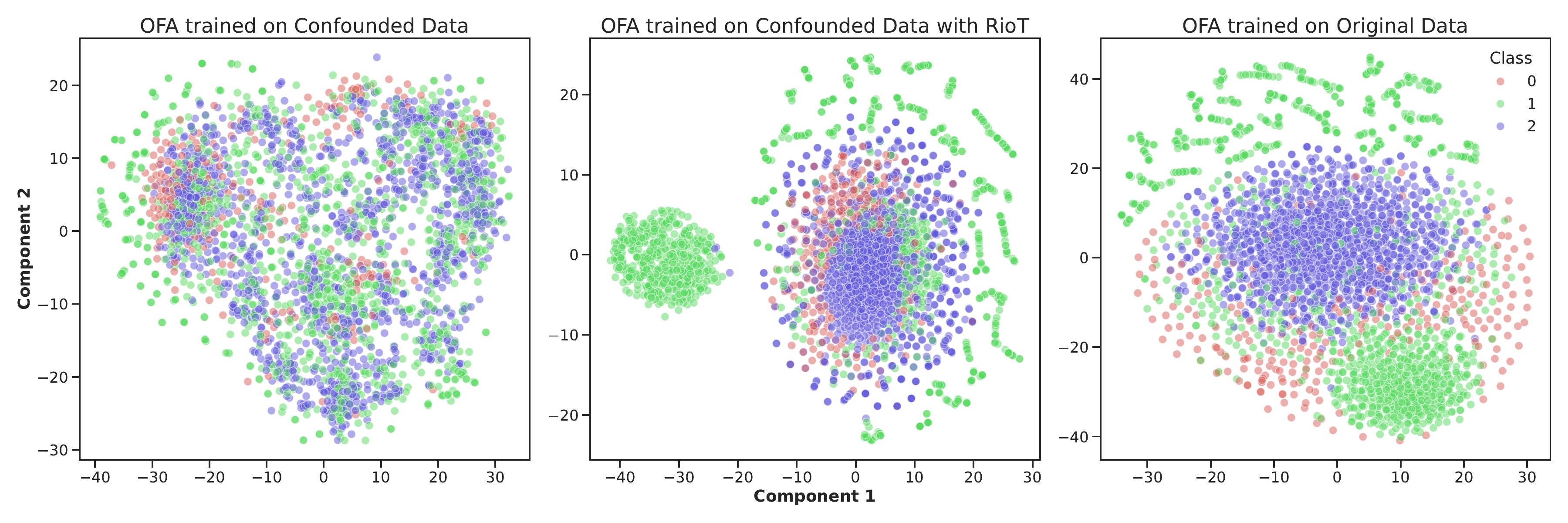}
    \caption{t-SNE plots of OFA encodings for Fault Detection A. The left plot shows that a model trained with shortcuts shows minimal class separation. The middle plot shows the same setup but after RioT regularization, while the far right plot shows an model without shortcuts with clear class separation. Both RioT-regularized and model without shortcuts exhibit similar structures, highlighting the effectiveness of RioT.}
    \label{fig:tsne}
\end{figure}
Lastly, we examine the inner workings of a model by analyzing its latent representations under various configurations.
In \autoref{fig:tsne}, t-SNE plots show OFA’s feature encodings on Fault Detection A in three settings: trained on shortcut data (left) with poor class separation; after RioT regularization (center), where structure and separation improve; and trained on clean data (right), yielding clear clusters. This reflects the scores of the models (\autoref{tab:xil_tsc_full}): the shortcut model reaches $\approx$50\% accuracy, whereas RioT boosts it to nearly 100\%, matching the reference scenario with clean data. This further demonstrates RioT's ability to mitigate shortcuts and restore robust performance.

\textbf{Limitations.}
A key aspect of \rrt is the incorporation of feedback. While this is a major advantage of \rrt, obtaining feedback can also present some challenges. Although we demonstrate that only a small fraction of annotated samples is needed, annotations remain essential. Moreover, like many interactive learning approaches, \rrt assumes accurate feedback, making it important to consider potential issues from inaccuracies in practical applications. To reduce the need of manual feedback, one could explore automated feedback strategies instead or alongside manual feedback \cite{stammer2024learning} (e.g., using an LLM to provide feedback or automated clustering of explanations to identify outliers). Such approaches may alleviate annotation costs but inevitably trade off some precision and can introduce new failure modes if the surrogate feedback is misaligned with task requirements.
Another drawback of \rrt is the increased training cost. Optimizing model explanations with gradient-based attributions requires computing a mixed-partial derivative. However, this can be efficiently handled using a Hessian-vector product, keeping the additional overhead manageable.

\section{Conclusion} \label{sec:conclusion}
In this work, we present \rrtfull(\rrt) a method to mitigate shortcuts in time series data with the help of feedback. By revising the model, \rrt significantly diminishes the influence of these factors, steering the model to align with the correct reasons. 
Using popular time series models on several controlled decoy datasets and the newly introduced, real-world dataset \data with naturally occurring shortcuts, showcases that SOTA models are indeed subject to shortcuts. Our results demonstrate that applying \rrt to these models can mitigate shortcuts in the data. Furthermore, we have unveiled that addressing solely the time domain is insufficient for fully steering the model toward the correct reasons. To overcome this, we extended our method to incorporate feedback in the frequency domain, offering an additional mechanism for reducing reliance on shortcuts.
Logical next steps are the extension of \rrt to multivariate time series and the integration of various explainer types. Furthermore, exploring the usage of adaptive feedback mechanisms could prove to be beneficial, in particular in the context of multiple simultaneous shortcuts. Beyond time series, the application of \rrt, especially \rrtf, can also allow for a more nuanced approach to shortcut mitigation in other modalities.

\begin{credits}
\subsubsection{\ackname} This work was partly funded by the Federal Ministry of Education and Research (BMBF) project “XEI” (FKZ 01IS24079B),  received funding by the EU project EXPLAIN, funded by the Federal Ministry of Education and Research (grant 01—S22030D). Additionally, it was funded by the project “The Adaptive Mind” from the Hessian Ministry of Science and the Arts (HMWK), the “ML2MT” project from the Volkswagen Stiftung, and the Priority Program (SPP) 2422 in the subproject “Optimization of active surface design of high-speed progressive tools using machine and deep learning algorithms” funded by the German Research Foundation (DFG). The latter also contributed the data for the P2S dataset. Furthermore, this work benefited from the HMWK project “The Third Wave of Artificial Intelligence – 3AI".

\subsubsection{\discintname} The authors have no competing interests to declare that are relevant to the content of this article.
\end{credits}

\bibliographystyle{splncs04}
\bibliography{references}

\newpage
\appendix
\FloatBarrier
\section{Appendix}

\subsection{Implementation and Experimental Details}\label{sec:app_model_details}

\subsubsection{Adaption of Integrated Gradients (IG)} 
A part of IG is a multiplication of the model gradient with the input itself, improving the explanation's quality \cite{shrikumarNotJustBlack2017}. However, this multiplication makes some implicit assumptions about the input format. In particular, it assumes that there are no inputs with negative values. Otherwise, multiplying the attribution score with a negative input would flip the attribution's sign, which is not desired. For images, this is unproblematic because they are always equal to or larger than zero. In time series, negative values can occur and normalization to make them all positive is not always suitable. To avoid this problem, we use only the input magnitude and not the input sign to compute the IG attributions.

\subsubsection{Computing Explanations}
To compute explanations with Integrated Gradients, we followed the common practice of using a baseline of zeros. The standard approach worked well in our experiments, so we did not explore other baseline choices in this work. For the implementation, we utilized the widely-used Captum\footnote{\url{https://github.com/pytorch/captum}} library, with patched \verb+captum._utils.gradient.compute_gradients+ function to allow for the propagation of the gradient with respect to the input to be propagated back into the parameters.

\subsubsection{Model Training and Hyperparameters}
To find suitable parameters for model training, we performed a hyperparameter search over batch size, learning rate, and the number of training epochs. We then used these parameters for all model trainings and evaluations, with and without \rrt. In addition, we selected suitable $\lambda$ values for \rrt with a hyperparameter selection on the respective validation sets. The exact values for the model training parameters and the $\lambda$ values can be found in the provided code\codenote.

To avoid model overfitting on the forecasting datasets, we performed shifted sampling with a window size of half the lookback window.

\subsubsection{Code}
For the experiments, we based our model implementations on the following repositories:
    
\begin{itemize}
    \item FCN: \url{https://github.com/qianlima-lab/time-series-ptms/blob/master/model/tsm_model.py}
    \item OFA: \url{https://github.com/DAMO-DI-ML/NeurIPS2023-One-Fits-All/}
    \item NBEATS: \url{https://github.com/unit8co/darts/blob/master/darts/models/forecasting/nbeats.py}
    \item TiDE: \url{https://github.com/unit8co/darts/blob/master/darts/models/forecasting/tide_model.py}
    \item PatchTST: \url{https://github.com/awslabs/gluonts/tree/dev/src/gluonts/torch/model/patch_tst}
\end{itemize}

All experiments were executed using our Python 3.11 and PyTorch, which is available in the provided code. To ensure reproducibility and consistency, we utilized Docker. Configurations and Python executables for all experiments are provided in the repository.

\subsubsection{Hardware}\label{sec:app_exp}
To conduct our experiments, we utilized single GPUs from Nvidia DGX2 machines equipped with A100-40G and A100-80G graphics processing units.
By maintaining a consistent hardware setup and a controlled software environment, we aimed to ensure the reliability and reproducibility of our experimental results.

\subsection{UCR Dataset selection}\label{sec:app_ucr}
We focused our evaluation on a subset of UCR datasets with a minimum size. Our selection process was as follows: First, we discarded all multivariate datasets, as we only considered univariate data in this paper. Then we removed all datasets with time series of different length or missing values. We further excluded all datasets of the category \textit{SIMULATED}, to avoid datasets which were synthetic from the beginning. We furthermore considered only datasets with less than 10 classes, as having a per-class shortcuts on more than 10 classes would result in a very high number of different shortcuts, which would probably rarely happen. Besides these criteria, we discarded all datasets with less than 1000 training samples or a per sample length of less than 100, to avoid the small datasets of UCR, which leads to the resulting four datasets: Fault Detection A, Ford A, Ford B and Sleep.

\subsection{Computational Costs of \rrt}\label{sec:app_cost}
\newcommand{\blockfontsize}{\large}
Training a model with RioT induces additional computational costs. The right-reason term requires computations of additional gradients. Given a model $f_\theta(x)$, parameterized by $\theta$ and input $x$, then computing the right reason loss with a gradient-based explanation method requires the computation of the mixed-partial derivative $\frac{\partial^2 f_\theta(x)}{\partial \theta \partial x}$, as a gradient-based explanation includes the derivative $\frac{\partial f_\theta(x)}{\partial x}$. While this mixed partial derivative is a second order derivative, this does not substantially increase the computational costs of our method for two main reasons. First, we are never explicitly materializing the Hessian matrix. Second, the second-order component of our loss can be formulated as a Hessian-vector product:
\begin{equation}
    \frac{\partial \mathcal{L}}{\partial \theta} = g + \frac{\lambda}{2} H_{\theta x}(e(x) - a(x))
\end{equation}
where $g = \frac{\partial \mathcal{L}_\mathrm{RA}}{\partial \theta}$ is the partial derivative of the right answer loss and if $H$ is the full joint Hessian matrix of the loss with respect to $\theta$ and $x$, then $H_{\theta x}$ is the sub-block of this matrix mapping $x$ into $\theta$ (cf.~\autoref{fig:sub_block}), given by $H_{\theta x} = \frac{\partial^2 f_\theta(x)}{\partial \theta \partial x}$.
Hessian-vector products are known to be fast to compute \cite{martensDeepLearningHessianfree2010}, enabling the right-reason loss computation to scale to large models and inputs.
\begin{figure}
    \centering
    \begin{tikzpicture}[scale=0.7]
      \draw[thick] (0,0) rectangle (4,4);
      \draw[thick] (0,2) -- (4,2);
      \draw[thick] (2,0) -- (2,4);

      \fill[blue!30] (2,2) rectangle (4,4);
      \draw[very thick, blue] (2,2) rectangle (4,4);

      \node at (1,3) {\blockfontsize$H_{\theta \theta}$};
      \node[blue] at (3,3) {\blockfontsize$H_{\theta x}$};
      \node at (1,1) {\blockfontsize$H_{x \theta}$};
      \node at (3,1) {\blockfontsize$H_{xx}$};

      \node[below] at (1,-0.1) {\blockfontsize$\theta$};
      \node[below] at (3,-0.2) {\blockfontsize$x$};
      \node[left] at (-0.1,1) {\blockfontsize$x$};
      \node[left] at (-0.1,3) {\blockfontsize$\theta$};

    \end{tikzpicture}
    \caption{Illustration of the Hessian matrix with its respective sub-blocks. The mapping from \( x \) into \( \theta \) is highlighted in blue.}
    \label{fig:sub_block}
\end{figure}

\subsection{Details on Shortcuts}\label{sec:conf_details}
\begin{figure}
    \centering
    \includegraphics[width=0.8\linewidth]{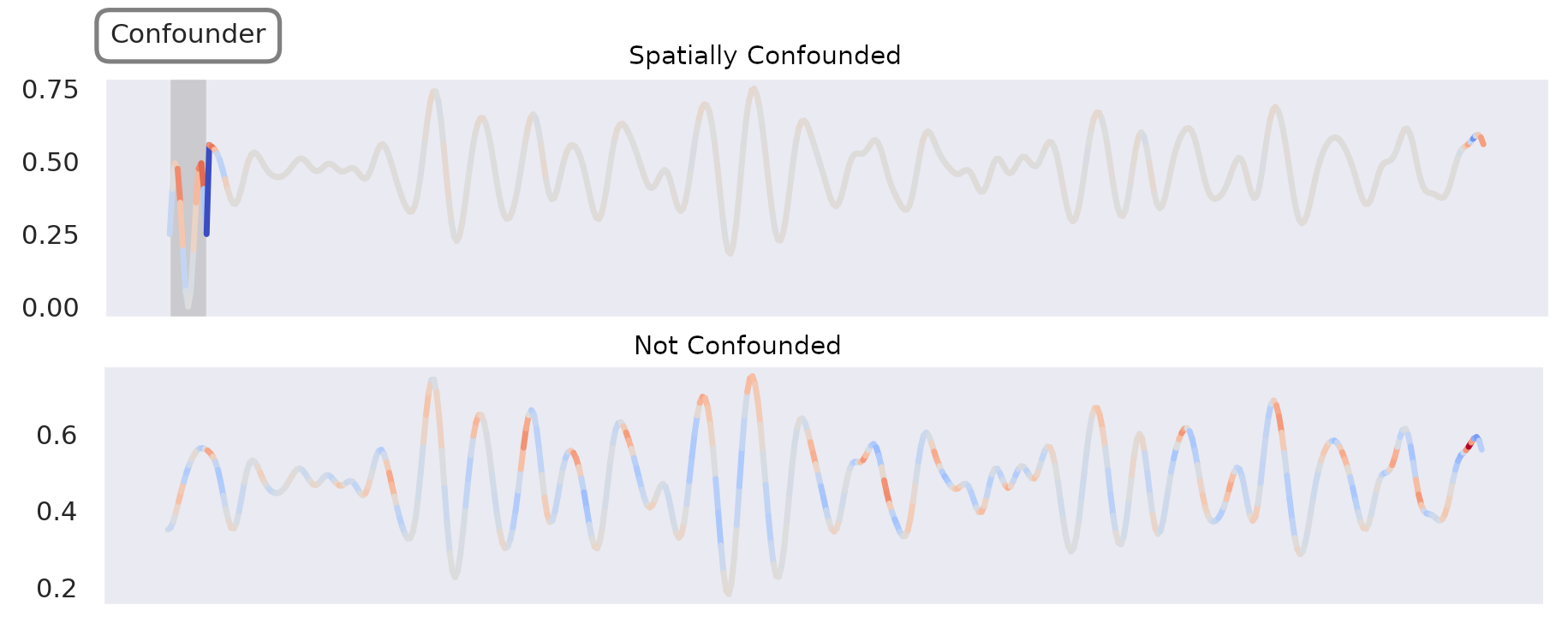}
    \caption{Example of the added spatial shortcut in the Fault Detection dataset.}
    \label{fig:spatial_conf_example}
\end{figure}

\begin{figure}
    \centering
    \includegraphics[width=0.55\linewidth]{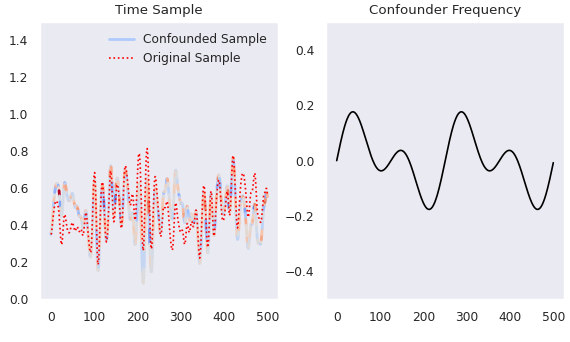}
    \caption{Example of an added frequency shortcut in the Fault Detection dataset.}
    \label{fig:freq_conf_example}
\end{figure}

In the datasets which are not \data, we added synthetic shortcuts to evaluate the effectiveness of shortcuts. Here, we provide some examples and more detailed descriptions of the decoy shortcuts. In \autoref{fig:spatial_conf_example} an example for a spatial decoy for classification is shown and in \autoref{fig:freq_conf_example} an examples for the frequency decoy for classification.

For forecasting datasets, spatial decoy are shortcuts that act as the actual solution to the forecasting problem. Periodically, data from the time series is copied back in time. This “back-copy” is a shortcut for the forecast, as it resembles the time steps of the forecasting window. Due to the windowed sampling from the time series, this shortcut occurs at every second sample. An exemplary visualization of this setting is shown in \autoref{fig:forecasting_spatial_conf}, with an exemplary lookback length of $9$, forecasting horizon of $3$ and window stride of $6$. This results in a shortcut in samples 1 and 3 (marked red) and overlapping in sample 2 (marked orange).

\begin{figure}
    \centering
    \def\scaling{0.63} %
    \begin{tikzpicture}[scale=\scaling]
        \def\boxwidth{1}
        \def\offset{1.5}
        \def\rowheight{0.9} %

        \node[anchor=west] at (-2,10) {\textbf{1. Sample}};
        \node[anchor=west] at (\offset-.1,10) {\textbf{Lookback}};
        \node[anchor=west] at (\offset+10.3,10) {\textbf{Horizon}};
        
        \node[anchor=west] at (-2,10-\rowheight) {No Shortcut:};
        \foreach \i/\t in {0/0, 1/1, 2/2, 3/3, 4/4, 5/5, 6/6, 7/7, 8/8} {
            \draw (\offset+\i*\boxwidth,10-\rowheight-0.25) rectangle (\offset+\i*\boxwidth+\boxwidth,10-\rowheight+0.25) node[pos=.5] {\t};
        }
        \foreach \i/\t in {9/9, 10/10, 11/11} {
            \draw (\i*\boxwidth+2*\offset,10-\rowheight-0.25) rectangle (\i*\boxwidth+\boxwidth+2*\offset,10-\rowheight+0.25) node[pos=.5] {\t};
        }

        \node[anchor=west] at (-2,10-2*\rowheight) {Shortcut:};
        \foreach \i/\t in {0/9, 1/10, 2/11} {
            \filldraw[fill=red!30, draw=red, thick] (\offset+\i*\boxwidth,10-2*\rowheight-0.25) rectangle (\offset+\i*\boxwidth+\boxwidth,10-2*\rowheight+0.25);
            \node at (\offset+\i*\boxwidth+\boxwidth/2,10-2*\rowheight) {\t};
        }
        \foreach \i/\t in {3/3, 4/4, 5/5, 6/6, 7/7, 8/8} {
            \draw (\offset+\i*\boxwidth,10-2*\rowheight-0.25) rectangle (\offset+\i*\boxwidth+\boxwidth,10-2*\rowheight+0.25);
            \node at (\offset+\i*\boxwidth+\boxwidth/2,10-2*\rowheight) {\t};
        }
        \foreach \i/\t in {9/9, 10/10, 11/11} {
            \draw[draw=red, thick] (\i*\boxwidth+2*\offset,10-2*\rowheight-0.25) rectangle (\i*\boxwidth+\boxwidth+2*\offset,10-2*\rowheight+0.25) node[pos=.5] {\t};
        }

        \node[anchor=west] at (-2,10-3*\rowheight) {Feedback:};
        \foreach \i/\t in {0/1, 1/1, 2/1, 3/0, 4/0, 5/0, 6/0, 7/0, 8/0} {
            \draw (\offset+\i*\boxwidth,10-3*\rowheight-0.25) rectangle (\offset+\i*\boxwidth+\boxwidth,10-3*\rowheight+0.25) node[pos=.5] {\t};
        }

        \node[anchor=west] at (-2,10-4*\rowheight) {\textbf{2. Sample}};
        
        \node[anchor=west] at (-2,10-5*\rowheight) {No Shortcut:};
        \foreach \i/\t in {0/6, 1/7, 2/8, 3/9, 4/10, 5/11, 6/12, 7/13, 8/14} {
            \draw (\offset+\i*\boxwidth,10-5*\rowheight-0.25) rectangle (\offset+\i*\boxwidth+\boxwidth,10-5*\rowheight+0.25) node[pos=.5] {\t};
        }
        \foreach \i/\t in {9/15, 10/16, 11/17} {
            \draw (\i*\boxwidth+2*\offset,10-5*\rowheight-0.25) rectangle (\i*\boxwidth+\boxwidth+2*\offset,10-5*\rowheight+0.25) node[pos=.5] {\t};
        }

        \node[anchor=west] at (-2,10-6*\rowheight) {Shortcut:};
        \foreach \i/\t in {0/6, 1/7, 2/8, 3/9, 4/10, 5/11} {
            \draw (\offset+\i*\boxwidth,10-6*\rowheight-0.25) rectangle (\offset+\i*\boxwidth+\boxwidth,10-6*\rowheight+0.25) node[pos=.5] {\t};
        }
        \foreach \i/\t in {6/21, 7/22, 8/23} {
            \filldraw[fill=orange!30, draw=orange, thick] (\offset+\i*\boxwidth,10-6*\rowheight-0.25) rectangle (\offset+\i*\boxwidth+\boxwidth,10-6*\rowheight+0.25) node[pos=.5] {\t};
        }
        \foreach \i/\t in {9/15, 10/16, 11/17} {
            \draw (\i*\boxwidth+2*\offset,10-6*\rowheight-0.25) rectangle (\i*\boxwidth+\boxwidth+2*\offset,10-6*\rowheight+0.25) node[pos=.5] {\t};
        }

        \node[anchor=west] at (-2,10-7*\rowheight) {Feedback:};
        \foreach \i/\t in {0/0, 1/0, 2/0, 3/0, 4/0, 5/0, 6/0, 7/0, 8/0} {
            \draw (\offset+\i*\boxwidth,10-7*\rowheight-0.25) rectangle (\offset+\i*\boxwidth+\boxwidth,10-7*\rowheight+0.25) node[pos=.5] {\t};
        }

        \node[anchor=west] at (-2,10-8*\rowheight) {\textbf{3. Sample}};
        
        \node[anchor=west] at (-2,10-9*\rowheight) {No Shortcut:};
        \foreach \i/\t in {0/12, 1/13, 2/14, 3/15, 4/16, 5/17, 6/18, 7/19, 8/20} {
            \draw (\offset+\i*\boxwidth,10-9*\rowheight-0.25) rectangle (\offset+\i*\boxwidth+\boxwidth,10-9*\rowheight+0.25) node[pos=.5] {\t};
        }
        \foreach \i/\t in {9/21, 10/22, 11/23} {
            \draw (\i*\boxwidth+2*\offset,10-9*\rowheight-0.25) rectangle (\i*\boxwidth+\boxwidth+2*\offset,10-9*\rowheight+0.25) node[pos=.5] {\t};
        }

        \node[anchor=west] at (-2,10-10*\rowheight) {Shortcut:};
        \foreach \i/\t in {0/21, 1/22, 2/23} {
            \filldraw[fill=red!30, draw=red, thick] (\offset+\i*\boxwidth,10-10*\rowheight-0.25) rectangle (\offset+\i*\boxwidth+\boxwidth,10-10*\rowheight+0.25);
            \node at (\offset+\i*\boxwidth+\boxwidth/2,10-10*\rowheight) {\t};
        }
        \foreach \i/\t in {3/15, 4/16, 5/17, 6/18, 7/19, 8/20} {
            \draw (\offset+\i*\boxwidth,10-10*\rowheight-0.25) rectangle (\offset+\i*\boxwidth+\boxwidth,10-10*\rowheight+0.25);
            \node at (\offset+\i*\boxwidth+\boxwidth/2,10-10*\rowheight) {\t};
        }
        \foreach \i/\t in {9/21, 10/22, 11/23} {
            \draw[draw=red, thick] (\i*\boxwidth+2*\offset,10-10*\rowheight-0.25) rectangle (\i*\boxwidth+\boxwidth+2*\offset,10-10*\rowheight+0.25) node[pos=.5] {\t};
        }

        \node[anchor=west] at (-2,10-11*\rowheight) {Feedback:};
        \foreach \i/\t in {0/1, 1/1, 2/1, 3/0, 4/0, 5/0, 6/0, 7/0, 8/0} {
            \draw (\offset+\i*\boxwidth,10-11*\rowheight-0.25) rectangle (\offset+\i*\boxwidth+\boxwidth,10-11*\rowheight+0.25) node[pos=.5] {\t};
        }

        \node[anchor=west] at (-2,10-13.5*\rowheight) {Shortcut:};
        \filldraw[fill=red!30, draw=red, thick] (\offset*0.5,10-13.5*\rowheight-0.25) rectangle (\offset*0.5+\boxwidth,10-13.5*\rowheight+0.25);
        
        \node[anchor=west] at (2,10-13.5*\rowheight) {Overlapping Shortcut:};
        \filldraw[fill=orange!30, draw=orange, thick] (6.25+\offset*0.5,10-13.5*\rowheight-0.25) rectangle (6.25+\offset*0.5+\boxwidth,10-13.5*\rowheight+0.25);
    \end{tikzpicture}
    \caption{Schematic overview of spatial decoys for the forecasting experiments.}
    \label{fig:forecasting_spatial_conf}
\end{figure}
\FloatBarrier
\subsection{Additional Experimental Results}
\label{sec:app_add_res}

\begin{table*}[ht]
\centering
\scriptsize
\caption{Feedback percentage for forecasting across all datasets, reported for the TiDE model. Corresponding to (test) results shown in \autoref{fig:scaling_exp}, a higher percentage indicates more feedback, lower is better.}
\label{tab:scaling_forecasting}
\vskip 0.1in
\begin{tabular}{llcccccc}
\toprule
Metric & Feedback & \multicolumn{2}{c}{ETTM1} & \multicolumn{2}{c}{Energy} & \multicolumn{2}{c}{Weather} \\
 &  & Spatial & Freq & Spatial & Freq & Spatial & Freq \\
\midrule
MAE ($\downarrow$) & 0\% & 0.54 \textpm \tiny0.01 & 0.74 \textpm \tiny0.06 & 0.85 \textpm \tiny0.01 & 0.53 \textpm \tiny0.07 & 0.29 \textpm \tiny0.01 & 0.49 \textpm \tiny0.09 \\
 & 5\% & 0.52 \textpm \tiny0.00 & 0.63 \textpm \tiny0.03 & 0.62 \textpm \tiny0.01 & \textbf{0.40} \textpm \tiny0.02 & 0.28 \textpm \tiny0.01 & 0.43 \textpm \tiny0.03 \\
 & 10\% & 0.52 \textpm \tiny0.00 & 0.63 \textpm \tiny0.03 & 0.61 \textpm \tiny0.01 & \textbf{0.40} \textpm \tiny0.02 & 0.27 \textpm \tiny0.01 & 0.43 \textpm \tiny0.03 \\
 & 25\% & 0.52 \textpm \tiny0.00 & 0.63 \textpm \tiny0.03 & 0.58 \textpm \tiny0.01 & 0.41 \textpm \tiny0.01 & 0.25 \textpm \tiny0.01 & 0.43 \textpm \tiny0.04 \\
 & 50\% & 0.52 \textpm \tiny0.00 & 0.63 \textpm \tiny0.03 & \textbf{0.57} \textpm \tiny0.01 & 0.41 \textpm \tiny0.01 & \textbf{0.24} \textpm \tiny0.01 & 0.44 \textpm \tiny0.05 \\
 & 75\% & 0.52 \textpm \tiny0.01 & 0.63 \textpm \tiny0.03 & \textbf{0.57} \textpm \tiny0.01 & 0.41 \textpm \tiny0.01 & \textbf{0.24} \textpm \tiny0.01 & 0.45 \textpm \tiny0.06 \\
 & 100\% & \textbf{0.51} \textpm \tiny0.01 & \textbf{0.60} \textpm \tiny0.05 & 0.58 \textpm \tiny0.01 & \textbf{0.40} \textpm \tiny0.03 & \textbf{0.24} \textpm \tiny0.01 & \textbf{0.41} \textpm \tiny0.02 \\
\midrule\midrule
MSE ($\downarrow$) & 0\% & 0.54 \textpm \tiny0.03 & 0.69 \textpm \tiny0.08 & 1.19 \textpm \tiny0.03 & 0.34 \textpm \tiny0.08 & 0.15 \textpm \tiny0.01 & 0.31 \textpm \tiny0.09 \\
 & 5\% & 0.54 \textpm \tiny0.01 & 0.52 \textpm \tiny0.03 & 0.60 \textpm \tiny0.02 & \textbf{0.20} \textpm \tiny0.01 & 0.14 \textpm \tiny0.01 & 0.24 \textpm \tiny0.02 \\
 & 10\% & 0.53 \textpm \tiny0.01 & 0.52 \textpm \tiny0.03 & 0.57 \textpm \tiny0.02 & \textbf{0.20} \textpm \tiny0.01 & 0.14 \textpm \tiny0.01 & 0.24 \textpm \tiny0.02 \\
 & 25\% & 0.53 \textpm \tiny0.01 & 0.52 \textpm \tiny0.03 & 0.53 \textpm \tiny0.02 & 0.22 \textpm \tiny0.01 & \textbf{0.11} \textpm \tiny0.01 & 0.24 \textpm \tiny0.03 \\
 & 50\% & 0.53 \textpm \tiny0.01 & 0.52 \textpm \tiny0.03 & \textbf{0.51} \textpm \tiny0.02 & 0.22 \textpm \tiny0.01 & \textbf{0.11} \textpm \tiny0.01 & 0.25 \textpm \tiny0.04 \\
 & 75\% & 0.52 \textpm \tiny0.01 & 0.51 \textpm \tiny0.03 & 0.52 \textpm \tiny0.02 & 0.22 \textpm \tiny0.01 & \textbf{0.11} \textpm \tiny0.01 & 0.26 \textpm \tiny0.05 \\
 & 100\% & \textbf{0.48} \textpm \tiny0.01 & \textbf{0.49} \textpm \tiny0.07 & 0.52 \textpm \tiny0.02 & 0.21 \textpm \tiny0.02 & \textbf{0.11} \textpm \tiny0.01 & \textbf{0.22} \textpm \tiny0.01 \\
\bottomrule
\end{tabular}

\end{table*}
\begin{table*}[ht]
\centering
\tiny
\caption{Feedback (FB) percentage for classification across all datasets, reported for the FCN model. Corresponding to results shown in \autoref{fig:scaling_exp}, a higher percentage indicates more feedback, higher is better.}
\label{tab:scaling_classification}
\vskip 0.1in
\begin{tabular}{lcccccccc}
\toprule
FB & \multicolumn{2}{c}{Fault Detection A (ACC $\uparrow$)} & \multicolumn{2}{c}{FordA (ACC $\uparrow$)} & \multicolumn{2}{c}{FordB (ACC $\uparrow$)} & \multicolumn{2}{c}{Sleep (ACC $\uparrow$)} \\
 & Spatial & Freq & Spatial & Freq & Spatial & Freq & Spatial & Freq \\
\midrule
0\% & 0.74 \textpm 0.06 & 0.87 \textpm 0.03 & 0.71 \textpm 0.08 & 0.73 \textpm 0.01 & 0.63 \textpm 0.03 & 0.60 \textpm 0.01 & 0.10 \textpm 0.03 & 0.27 \textpm 0.02 \\
5\% & 0.88 \textpm 0.00 & 0.88 \textpm 0.01 & 0.81 \textpm 0.03 & 0.80 \textpm 0.03 & 0.66 \textpm 0.03 & \textbf{0.66} \textpm 0.02 & 0.53 \textpm 0.03 & \textbf{0.49} \textpm 0.00 \\
10\% & 0.89 \textpm 0.02 & 0.89 \textpm 0.01 & 0.82 \textpm 0.04 & 0.79 \textpm 0.02 & 0.66 \textpm 0.03 & 0.64 \textpm 0.03 & 0.48 \textpm 0.09 & 0.48 \textpm 0.02 \\
25\% & 0.92 \textpm 0.01 & 0.89 \textpm 0.01 & 0.83 \textpm 0.02 & 0.78 \textpm 0.01 & 0.67 \textpm 0.02 & 0.65 \textpm 0.01 & 0.49 \textpm 0.08 & 0.42 \textpm 0.08 \\
50\% & \textbf{0.95} \textpm 0.01 & 0.88 \textpm 0.01 & 0.82 \textpm 0.03 & 0.81 \textpm 0.05 & 0.67 \textpm 0.02 & 0.65 \textpm 0.00 & \textbf{0.55} \textpm  0.03 & 0.44 \textpm 0.07 \\
75\% & \textbf{0.95} \textpm 0.01 & 0.88 \textpm 0.01 & 0.81 \textpm 0.03 & 0.80 \textpm 0.04 & 0.65 \textpm 0.03 & 0.64 \textpm 0.00 & 0.54 \textpm 0.04 & 0.44 \textpm 0.07 \\
100\% & 0.93 \textpm 0.03 & \textbf{0.90} \textpm 0.03 & \textbf{0.84} \textpm 0.02 & \textbf{0.83} \textpm 0.02 & \textbf{0.68} \textpm 0.02 & 0.65 \textpm 0.01 & 0.54 \textpm 0.05 & 0.45 \textpm 0.07 \\
\bottomrule
\end{tabular}

\end{table*}
\begin{table*}[th]
\centering
\tiny
\caption{\textbf{\rrt can successfully overcome shortcuts in time series forecasting.}
MAE values on the training set with and the test set without shortcuts. \textit{No Shortcut} is the ideal scenario where the model is not affected by any shortcut.}
\label{tab:xil_tsf_full_mae}
\vskip 0.1in
\setlength{\tabcolsep}{5.1pt}
\begin{tabular}{lll@{\hspace{1pt}}cl@{\hspace{1pt}}cl@{\hspace{1pt}}c}
    \toprule
    Model & Config (MAE $\downarrow$) & \multicolumn{2}{c}{ETTM1} & \multicolumn{2}{c}{Energy} & \multicolumn{2}{c}{Weather} \\
     &  & Train & \makebox[5pt][r]{Test} & Train & \makebox[5pt][r]{Test} & Train & \makebox[5pt][r]{Test} \\
    \midrule
    NBEATS & No Shortcut & 0.39 \textpm \tiny 0.01 & \multicolumn{1}{r}{0.48 \textpm \tiny 0.01} & 0.44 \textpm \tiny 0.02 & \multicolumn{1}{r}{0.38 \textpm \tiny 0.01} & 0.21 \textpm \tiny 0.01 & \multicolumn{1}{r}{0.12 \textpm \tiny 0.01} \\
    \cmidrule{2-8}\morecmidrules\cmidrule{2-8}
     & Base\textsubscript{sp} & \textbf{0.34} \textpm \tiny 0.01 & \multicolumn{1}{r}{0.54 \textpm \tiny 0.01} & \textbf{0.44} \textpm \tiny 0.03 & \multicolumn{1}{r}{0.77 \textpm \tiny 0.01} & \textbf{0.21} \textpm \tiny 0.01 & \multicolumn{1}{r}{0.30 \textpm \tiny 0.04} \\
     & + RioT\textsubscript{sp} & 0.40 \textpm \tiny 0.01 & \multicolumn{1}{r}{\textbf{0.52} \textpm \tiny 0.01} & 0.53 \textpm \tiny 0.02 & \multicolumn{1}{r}{\textbf{0.62} \textpm \tiny 0.01} & 0.23 \textpm \tiny 0.01 & \multicolumn{1}{r}{\textbf{0.22} \textpm \tiny 0.01} \\
    \cmidrule{2-8}
     & Base\textsubscript{freq} & \textbf{0.39} \textpm \tiny 0.01 & \multicolumn{1}{r}{0.47 \textpm \tiny 0.01} & \textbf{0.45} \textpm \tiny 0.03 & \multicolumn{1}{r}{0.45 \textpm \tiny 0.03} & \textbf{0.21} \textpm \tiny 0.03 & \multicolumn{1}{r}{0.45 \textpm \tiny 0.06} \\
     & + RioT\textsubscript{freq} & 0.40 \textpm \tiny 0.01 & \multicolumn{1}{r}{\textbf{0.47} \textpm \tiny 0.01} & \textbf{0.45} \textpm \tiny 0.03 & \multicolumn{1}{r}{\textbf{0.44} \textpm \tiny 0.02} & 0.59 \textpm \tiny 0.22 & \multicolumn{1}{r}{\textbf{0.39} \textpm \tiny 0.01} \\
    \midrule
    PatchTST & No Shortcut & 0.50 \textpm \tiny 0.01 & \multicolumn{1}{r}{0.49 \textpm \tiny 0.01} & 0.39 \textpm \tiny 0.00 & \multicolumn{1}{r}{0.38 \textpm \tiny 0.01} & 0.38 \textpm \tiny 0.03 & \multicolumn{1}{r}{0.18 \textpm \tiny 0.00} \\
    \cmidrule{2-8}\morecmidrules\cmidrule{2-8}
     & Base\textsubscript{sp} & \textbf{0.46} \textpm \tiny 0.00 & \multicolumn{1}{r}{0.53 \textpm \tiny 0.01} & \textbf{0.41} \textpm \tiny 0.00 & \multicolumn{1}{r}{0.78 \textpm \tiny 0.01} & \textbf{0.32} \textpm \tiny 0.04 & \multicolumn{1}{r}{0.33 \textpm \tiny 0.00} \\
     & + RioT\textsubscript{sp} & \textbf{0.46} \textpm \tiny 0.01 & \multicolumn{1}{r}{\textbf{0.52} \textpm \tiny 0.01} & 0.51 \textpm \tiny 0.00 & \multicolumn{1}{r}{\textbf{0.53} \textpm \tiny 0.01} & 0.54 \textpm \tiny 0.12 & \multicolumn{1}{r}{\textbf{0.28} \textpm \tiny 0.00} \\
    \cmidrule{2-8}
     & Base\textsubscript{freq} & \textbf{0.53} \textpm \tiny 0.01 & \multicolumn{1}{r}{0.81 \textpm \tiny 0.07} & \textbf{0.15} \textpm \tiny 0.00 & \multicolumn{1}{r}{0.64 \textpm \tiny 0.03} & \textbf{0.58} \textpm \tiny 0.03 & \multicolumn{1}{r}{0.41 \textpm \tiny 0.05} \\
     & + RioT\textsubscript{freq} & 0.92 \textpm \tiny 0.05 & \multicolumn{1}{r}{\textbf{0.80} \textpm \tiny 0.02} & 0.97 \textpm \tiny 0.86 & \multicolumn{1}{r}{\textbf{0.57} \textpm \tiny 0.02} & 0.65 \textpm \tiny 0.01 & \multicolumn{1}{r}{\textbf{0.40} \textpm \tiny 0.01} \\
    \midrule
    TiDE & No Shortcut & 0.36 \textpm \tiny 0.01 & \multicolumn{1}{r}{0.48 \textpm \tiny 0.01} & 0.40 \textpm \tiny 0.01 & \multicolumn{1}{r}{0.38 \textpm \tiny 0.02} & 0.36 \textpm \tiny 0.02 & \multicolumn{1}{r}{0.13 \textpm \tiny 0.00} \\
    \cmidrule{2-8}\morecmidrules\cmidrule{2-8}
     & Base\textsubscript{sp} & \textbf{0.32} \textpm \tiny 0.01 & \multicolumn{1}{r}{0.54 \textpm \tiny 0.01} & \textbf{0.40} \textpm \tiny 0.01 & \multicolumn{1}{r}{0.85 \textpm \tiny 0.01} & \textbf{0.32} \textpm \tiny 0.03 & \multicolumn{1}{r}{0.29 \textpm \tiny 0.01} \\
     & + RioT\textsubscript{sp} & 0.34 \textpm \tiny 0.01 & \multicolumn{1}{r}{\textbf{0.51} \textpm \tiny 0.01} & 0.57 \textpm \tiny 0.01 & \multicolumn{1}{r}{\textbf{0.58} \textpm \tiny 0.01} & 0.35 \textpm \tiny 0.03 & \multicolumn{1}{r}{\textbf{0.24} \textpm \tiny 0.01} \\
    \cmidrule{2-8}
     & Base\textsubscript{freq} & \textbf{0.18} \textpm \tiny 0.01 & \multicolumn{1}{r}{0.74 \textpm \tiny 0.06} & \textbf{0.18} \textpm \tiny 0.01 & \multicolumn{1}{r}{0.53 \textpm \tiny 0.07} & \textbf{0.65} \textpm \tiny 0.05 & \multicolumn{1}{r}{0.49 \textpm \tiny 0.09} \\
     & + RioT\textsubscript{freq} & 0.19 \textpm \tiny 0.01 & \multicolumn{1}{r}{\textbf{0.60} \textpm \tiny 0.05} & \textbf{0.18} \textpm \tiny 0.01 & \multicolumn{1}{r}{\textbf{0.40} \textpm \tiny 0.03} & 0.79 \textpm \tiny 0.16 & \multicolumn{1}{r}{\textbf{0.41} \textpm \tiny 0.02} \\
    \bottomrule
    \end{tabular}

\end{table*}
\begin{table}
    \caption{\textbf{\rrt can combine spatial and frequency feedback.} 
    MAE results when applying feedback in time and frequency with \rrt. Combining both feedback domains is superior to feedback on only one of the domains. 
    \textit{No Shortcut} values represent the ideal scenario when the model is not affected by any shortcut (mean and std over 5 runs).
    }
\label{tab:combined_xil_mae}
\vskip 0.1in
\small
\centering
\begin{tabular}{lcc}
\toprule
Energy (MAE $\downarrow$) & Train & \text{\ \ \ Test} \\
\midrule \midrule
TiDE\textsubscript{No Shortcut} & 0.40 \textpm \scriptsize{0.01} & \multicolumn{1}{r}{0.38 \textpm \scriptsize{0.02}} \\
\midrule \midrule
TiDE\textsubscript{freq,sp} & \textbf{0.30} \textpm \scriptsize{0.01} & \multicolumn{1}{r}{0.70 \textpm \scriptsize{0.02}} \\
TiDE\textsubscript{freq,sp} + \rrts & 0.34 \textpm \scriptsize{0.01} & \multicolumn{1}{r}{0.64 \textpm \scriptsize{0.01}} \\
TiDE\textsubscript{freq,sp} + \rrtf & 0.36 \textpm \scriptsize{0.01} & \multicolumn{1}{r}{0.60 \textpm \scriptsize{0.01}} \\
TiDE\textsubscript{freq,sp} + \rrtfs & 0.39 \textpm \scriptsize{0.01} & \textbf{0.55} \textpm \scriptsize{0.01} \\
\bottomrule
\end{tabular}
\end{table}

This section provides further insights into our experiments, covering both forecasting and classification tasks. Specifically, it showcases performance through various metrics such as MAE, MSE, and accuracy, qualitative insights about the influence of shortcuts, and explores different feedback configurations.

\subsubsection{Feedback Generalization}
\autoref{tab:scaling_classification} and \autoref{tab:scaling_forecasting} detail provided feedback percentages for forecasting and classification across all datasets, respectively. These tables report the performance of the TIDE and FCN models, highlighting how different levels of feedback impact model outcomes on various datasets. \autoref{tab:scaling_forecasting} focuses on MAE and MSE for forecasting, while \autoref{tab:scaling_classification} presents ACC for classification.

\subsubsection{Removing Shortcuts for Time Series Forecasting}
\autoref{tab:xil_tsf_full_mae} reports the MAE results for our forecasting experiment across different models, datasets and configurations. It emphasizes how well each model performs on both the training set with and test set without shortcuts. Both setups are shown with and without applying \rrt. The \textit{No Shortcut} configuration representing the ideal scenario where the model is unaffected by any shortcuts.

\subsubsection{Removing Multiple Shortcuts at Once} 
\autoref{tab:combined_xil_mae} reports the MAE values and illustrates the effectiveness of combining spatial and frequency feedback via RioT for the TiDE model. The results demonstrate significant improvements in forecasting accuracy when integrating both feedback domains compared to using them separately.

\subsubsection{Early Stopping as Shortcut Mitigation Baseline}
In this experiment, we compare the performance of a model with RioT to a model regularized via early stopping (which is decided on an validation set without shortcuts). In that, we stop model training if there are no improvements in the validation set for several epochs in the hope that it thus does not overfit to the shortcut. The results are presented in \autoref{tab:early_stopping_classification} for classification and \autoref{tab:early_stopping_forecasting} for forecasting.  We can observe that early stopping can help in some instances to achieve performances similar to \rrt (e.g. PatchTST with a frequency shortcut or FCN with a spatial shortcut). However, for the majority of cases the performance with early stopping is substantially lower than the performance with \rrt, signaling that early stopping alone is not a sufficient approach to overcome shortcuts.

\begin{table}[ht]
    \centering
    \caption{Early stopping on classification datasets.}
    \begin{tabular}{l cc cc}
    \toprule
         Config & \multicolumn{2}{c}{FCN} & \multicolumn{2}{c}{OFA} \\ 
                & Train & Test & Train & Test \\ \midrule
 No Shortcut   & 0.99 \textpm \scriptsize{0.00} & 0.99 \textpm \scriptsize{0.00} & 1.00 \textpm \scriptsize{0.00} & 0.98 \textpm \scriptsize{0.02} \\ 
\midrule \midrule
 Base\textsubscript{sp} & \textbf{1.00} \textpm \scriptsize{0.00} & 0.74 \textpm \scriptsize{0.06} & \textbf{1.00} \textpm \scriptsize{0.00} & 0.53 \textpm \scriptsize{0.02} \\
 $\text{+RioT}_\text{sp}$ & 0.98 \textpm \scriptsize{0.01} & \textbf{0.93} \textpm \scriptsize{0.03} & 0.96 \textpm \scriptsize{0.08} & \textbf{0.98} \textpm \scriptsize{0.01} \\
 $\text{+ES}_\text{sp}$   & 0.87 \textpm \scriptsize{0.01} & 0.91 \textpm \scriptsize{0.03} & 0.69 \textpm \scriptsize{0.02} & 0.67 \textpm \scriptsize{0.04} \\ 
 \midrule
 Base\textsubscript{freq} & \textbf{0.98} \textpm \scriptsize{0.01} & 0.87 \textpm \scriptsize{0.03} & \textbf{1.00} \textpm \scriptsize{0.00} & 0.72 \textpm \scriptsize{0.02} \\
 $\text{+RioT}_\text{freq}$ & 0.94 \textpm \scriptsize{0.00} & \textbf{0.90} \textpm \scriptsize{0.03} & 0.96 \textpm \scriptsize{0.02} & \textbf{0.98} \textpm \scriptsize{0.02} \\
 $\text{+ES}_\text{freq}$   & 0.83 \textpm \scriptsize{0.01} & 0.86 \textpm \scriptsize{0.02} & 0.81 \textpm \scriptsize{0.01} & 0.75 \textpm \scriptsize{0.02} \\
         \bottomrule
    \end{tabular}
    \label{tab:early_stopping_classification}
\end{table}

\begin{table}[ht]
    \centering
    \caption{Early stopping on forecasting datasets.}
    \begin{tabular}{l cc cc}
    \toprule
         Config & \multicolumn{2}{c}{PatchTST} & \multicolumn{2}{c}{TiDE} \\ 
                & Train & Test & Train & Test \\ \midrule
No Shortcut   & 0.26 \textpm \scriptsize{0.01} & 0.23 \textpm \scriptsize{  0.00} & 0.27 \textpm \scriptsize{  0.01} & 0.26 \textpm \scriptsize{  0.02} \\ 
\midrule \midrule
Base\textsubscript{sp} & \textbf{0.29} \textpm \scriptsize{  0.01} & 0.96 \textpm \scriptsize{  0.03} & \textbf{0.28} \textpm \scriptsize{  0.01} & 1.19 \textpm \scriptsize{  0.03} \\
$\text{+RioT}_\text{sp}$ & 0.44 \textpm \scriptsize{  0.00} & \textbf{0.45} \textpm \scriptsize{  0.01} & 0.53 \textpm \scriptsize{  0.02} & \textbf{0.52} \textpm \scriptsize{  0.02} \\
$\text{+ES}_\text{sp}$ & 0.48 \textpm \scriptsize{  0.05} & 0.68 \textpm \scriptsize{  0.03} & 1.20 \textpm \scriptsize{  0.25} & 0.81 \textpm \scriptsize{  0.08} \\ 
\midrule
Base\textsubscript{freq} & \textbf{0.04} \textpm \scriptsize{  0.00} & 0.53 \textpm \scriptsize{  0.05} & \textbf{0.07} \textpm \scriptsize{  0.01} & 0.34 \textpm \scriptsize{  0.08} \\
$\text{+RioT}_\text{freq}$ & 0.71 \textpm \scriptsize{  0.10} & \textbf{0.38} \textpm \scriptsize{  0.06} & \textbf{0.07} \textpm \scriptsize{  0.01} & \textbf{0.21} \textpm \scriptsize{  0.02} \\
$\text{+ES}_\text{freq}$ & 0.48 \textpm \scriptsize{  0.09} & 0.49 \textpm \scriptsize{  0.08} & 0.21 \textpm \scriptsize{  0.08} & 0.36 \textpm \scriptsize{  0.09} \\ 
         \bottomrule
    \end{tabular}
    \label{tab:early_stopping_forecasting}
\end{table}

\subsubsection{Incorporating Alternative Explanations based on Shapelets} 
Attribution-based methods are commonly used to provide explanations at the individual time-step level. However, such explanations can often appear fuzzy or ambiguous, leading to potential misinterpretations and complicating feedback. Humans naturally perceive and interpret shapes or patterns more intuitively than isolated time steps. To leverage this strength, we explored an alternative approach by extracting shapelets \cite{yeTimeSeriesShapelets2011} from one of our benchmark datasets. In doing so, we demonstrate that our method's effectiveness is not limited to deep learning models; it can also address shortcut learning in simpler, interpretable models based on shapelets.

\begin{figure}[ht]
    \includegraphics[width=1\linewidth]{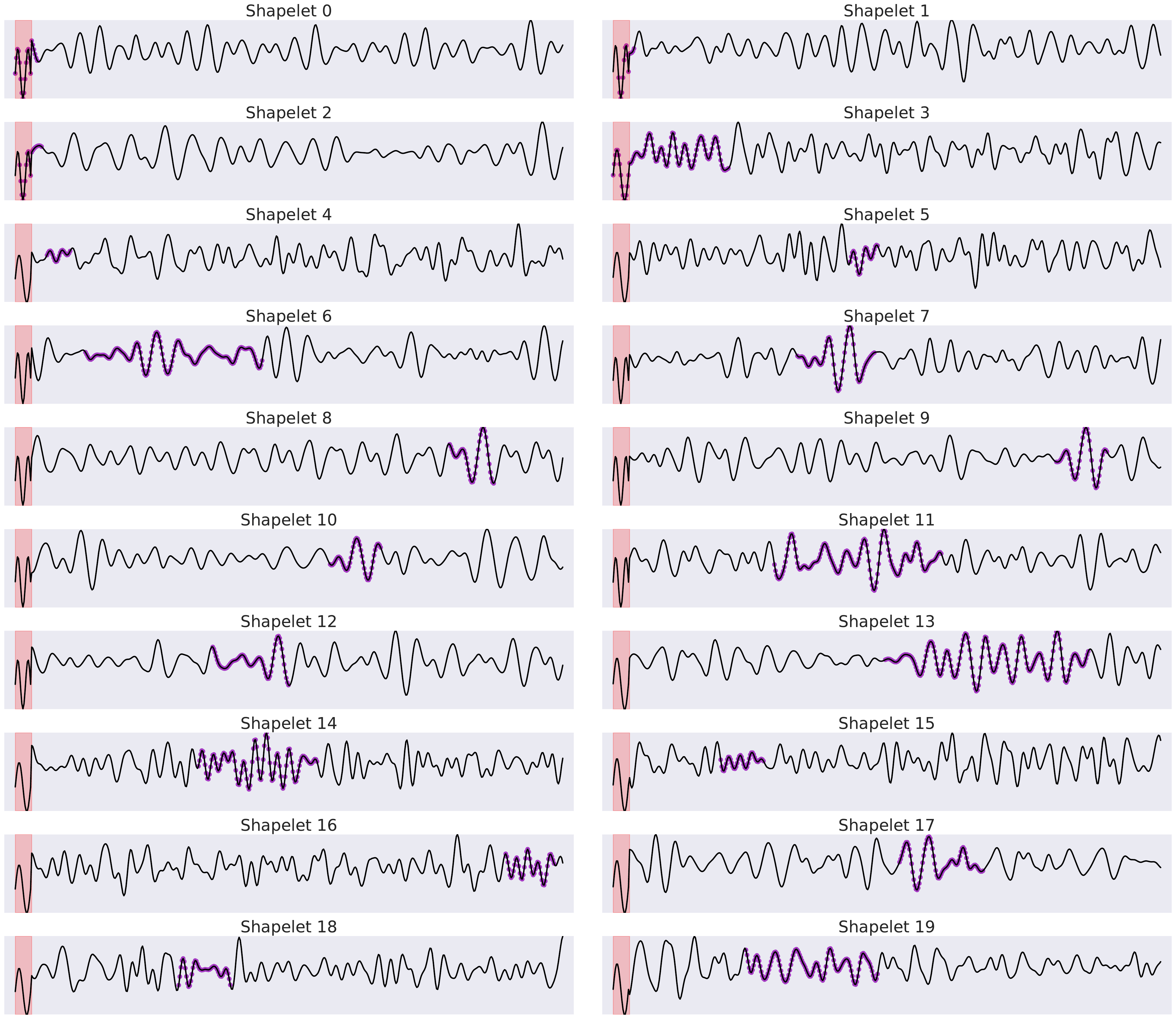}
    \caption{Computed shapelets of FordA. The purple segments represent the discovered shapelets overlaid on their corresponding training samples. The red region marks the included shortcut, and shapelets 0–3 explicitly model this shortcut.}
    \label{fig:shaplets}
\end{figure}

Specifically, we extracted shapelets (cf. \autoref{fig:shaplets}) from the dataset, trained a classifier based on these shapelets, and analyzed their attributions. Our observations revealed that the identified shortcut was captured within the activated shapelets. Crucially, we showed that our method (RioT) successfully mitigates this shortcut without compromising the activation of shapelets that represent meaningful features. 

We observe that in the baseline model, the first four shapelets—which cover the shortcut (highlighted in red)—are predominantly activated. In contrast, with RioT, the activation shifts away from the shapelets containing the shortcut. This observation is further quantitatively outlined in \autoref{tab:shaplet_results}, which summarizes the accuracy scores and aligns with our previous experimental results.

\begin{table}[h]
    \centering
    \caption{Accuracy on FordA when solving the task based on shapelets. }
    \begin{tabular}{lcc}
        \toprule
        Config & Train & Test\\
        \midrule
        No Shortcut & 0.90 \textpm \scriptsize{0.00} & 0.89 \textpm \scriptsize{0.00} \\
        \midrule \midrule
        Shapelets & \textbf{0.99} \textpm \scriptsize{0.00} & 0.72 \textpm \scriptsize{0.02} \\ 
        +\rrts & 0.92 \textpm \scriptsize{0.00} & \textbf{0.82} \textpm \scriptsize{0.00} \\
        \bottomrule
    \end{tabular}
    \label{tab:shaplet_results}
\end{table}

\FloatBarrier

\section{Dataset from a High-speed Progressive Tool with Natural Shortcuts} \label{sec:app_dataset}
The presence of shortcuts is a common challenge in practical settings, affecting models in diverse ways. As the research community strives to identify and mitigate these issues, it becomes imperative to test our methodologies on datasets that mirror the complexities encountered in actual applications. However, for the time domain, datasets with explicitly labeled shortcuts are not present, highlighting the challenge of assessing model performance against the complex nature of practical shortcuts.

To bridge this gap, we introduce \data, a dataset that represents a significant step forward by featuring explicitly identified shortcuts. This dataset originates from experimental work on a production line for deep-drawn sheet metal parts, employing a progressive die on a high-speed press. The sections below detail the experimental approach and the process of data collection.

\subsection{Real-World setting}
\begin{figure}[ht]
    \centering
    \includegraphics[width=1\columnwidth]{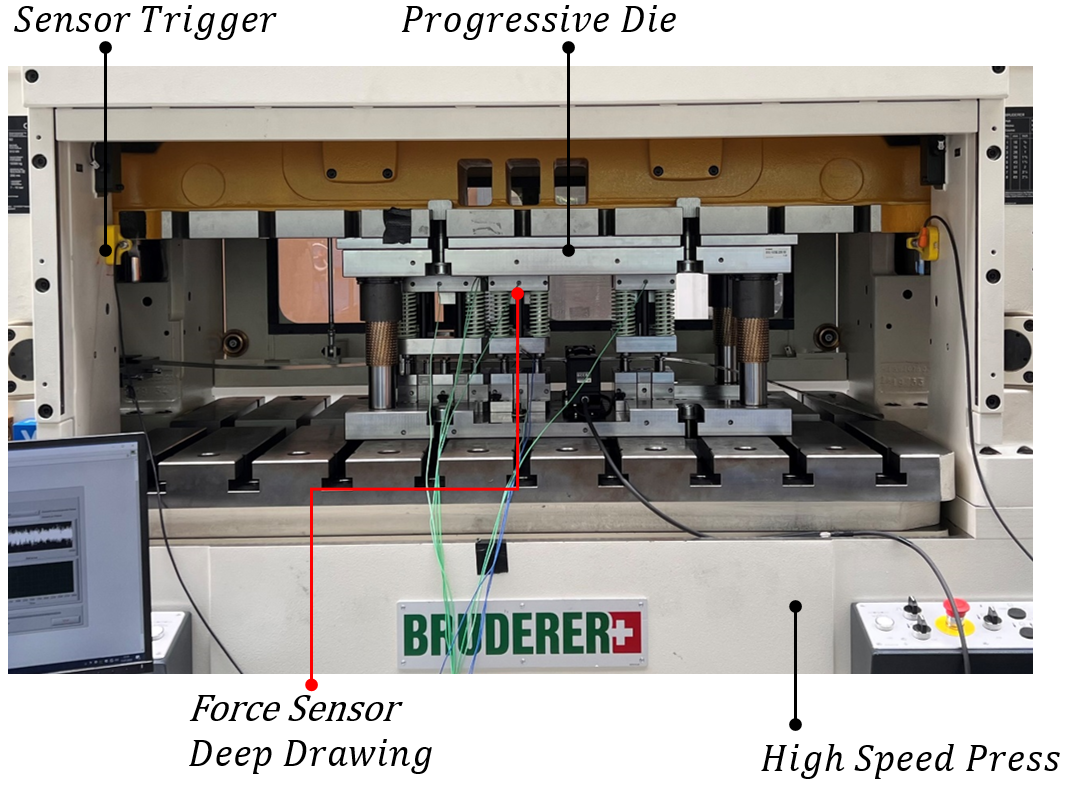}
    \caption{Experimental setup with high-speed press and tool as well as trigger for stroke-by-stroke recording of the data}
    \label{fig:ExpSetup}
\end{figure}

The production of parts in multiple progressive forming stages using stamping, deep drawing and bending operations with progressive dies is generally one of the most economically significant manufacturing processes in the sheet metal working industry and enables the production of complex parts on short process routes with consistent quality. For the tests, a four-stage progressive die was used on a Bruderer BSTA 810-145 high-speed press with varied stroke speed. The strip material to be processed is fed into the progressive die by a BSV300 servo feed unit, linked to the cycle of the press, in the stroke movement while the tools are not engaged. The part to be produced remains permanently connected to the sheet strip throughout the entire production run. The stroke height of the tool is 63 mm and the material feed per stroke is 60 mm. The experimental setup with the progressive die set up on the high-speed press is shown in \autoref{fig:ExpSetup}.

\begin{figure}[ht]
    \centering
    \includegraphics[width=1\columnwidth]{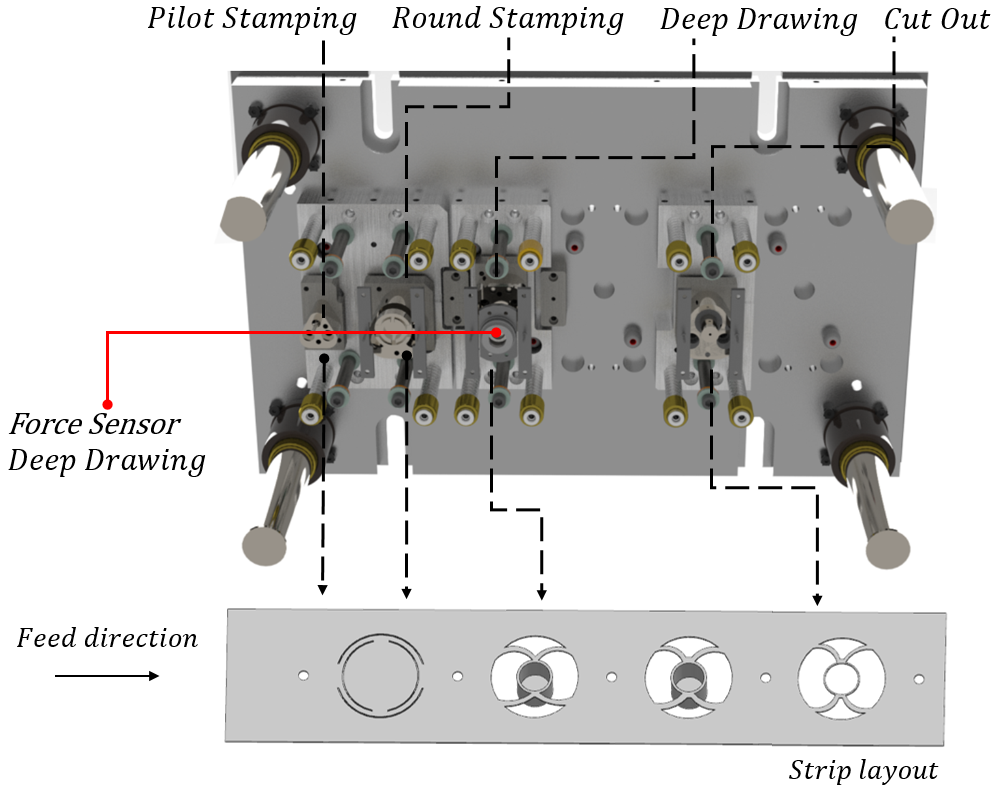}
    \caption{Upper tool unfolded and the forming stages associated with the respective steps on the passing sheet metal strip as well as the positions of the piezoelectric force sensors.}
    \label{fig:Process}
\end{figure}

The four stages include a pilot punching stage, a round stamping stage, deep drawing and a cut-out stage. In the first stage, a 3 mm hole is punched in the metal strip. This hole is used by guide pins in the subsequent stages to position the metal strip. During the stroke movement, the pilot pin always engages in the pilot hole first, thus ensuring the positioning accuracy of the components. In the subsequent stage, a circular blank is cut into the sheet metal strip. This is necessary so that the part can be deep-drawn directly from the sheet metal strip. This is a round geometry that forms small arms in the subsequent deep-drawing step that hold the component on the metal strip. In the final stage, the component is then separated from the sheet metal strip and the process cycle is completed. The respective process steps are performed simultaneously so that each stage carries out its respective process with each stroke and therefore a part is produced with each stroke. \autoref{fig:Process} shows the upper tool unfolded and the forming stages associated with the respective steps on the continuous sheet metal strip.

\subsection{Data collection}
An indirect piezoelectric force sensor (Kistler 9240A) was integrated into the upper mould mounting plate of the deep-drawing stage for data acquisition. The sensor is located directly above the punch and records not only the indirect process force but also the blank holder forces which are applied by spring assemblies between the upper mounting plate and the blank holder plate. The data is recorded at a sampling frequency of 25 kHz. The material used is DC04 with a width of 50 mm and a thickness of 0.5 mm. The voltage signals from the sensors are digitised using a "CompactRIO" (NI cRIO 9047) with integrated NI 9215 measuring module (analogue voltage input $\pm$ 10 V). Data recording is started via an inductive proximity switch when the press ram passes below a defined stroke height during the stroke movement and is stopped again as it passes the inductive proximity switch during the return stroke movement. Due to the varying process speed caused by the stroke speeds that have been set, the recorded time series have a different number of data points. Further, there are slight variations in the length of the time series withing one experiment. For this reason, all time series are interpolated to a length of 4096 data points and we discard any time series that deviate considerably from the mean length of a run (i.e., threshold of 3). A total of 12 series of experiments, shown in \autoref{tab:me_experiments}, were carried out with production rates from 80 to 225 spm. To simulate a defect, the spring hardness of the blank holder was manipulated in the test series that were marked as \textit{defect}. The manipulated experiments result in the component bursting and tearing during production. In a real production environment, this would lead directly to the parts being rejected.

\subsection{Data characteristics}
\begin{figure}[ht]
    \centering
    \includegraphics[width=0.7\columnwidth]{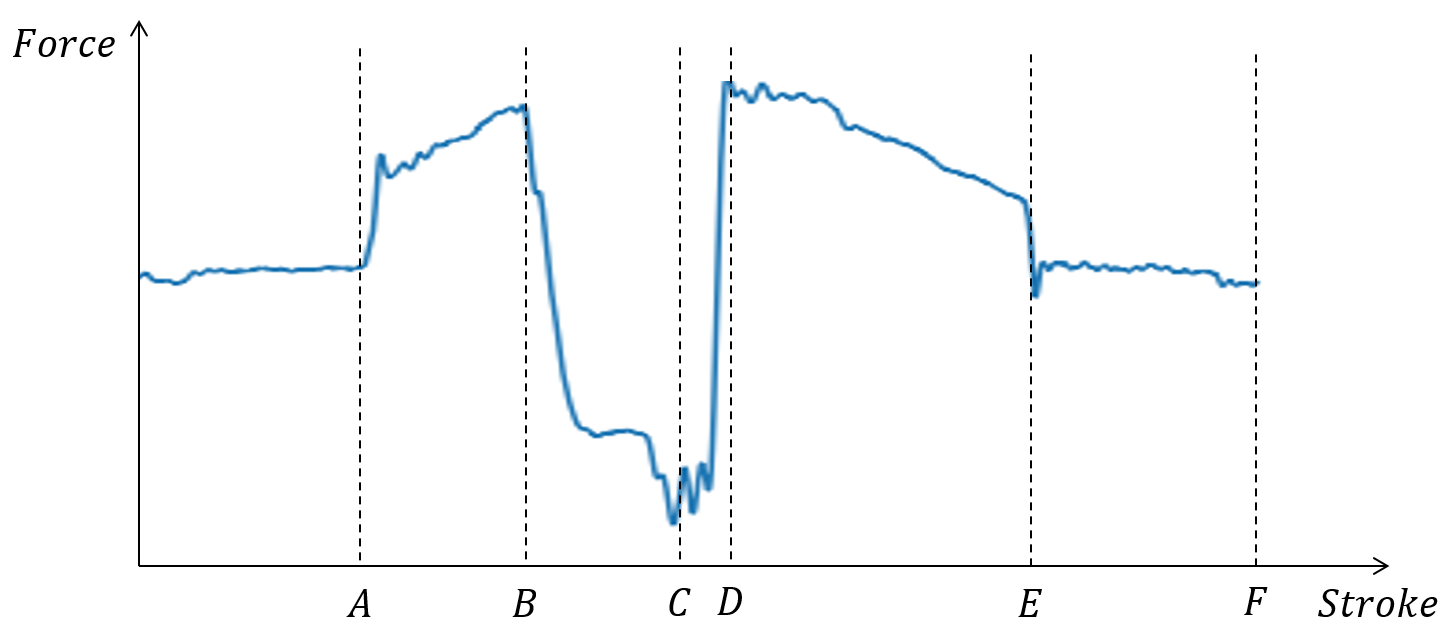}
    \caption{Force curve for one stroke. A) set down downholder plate B) start of deep drawing C) bottom dead centre D) deep drawing process completed E) downholder plates lift off F) measurement stops.}
    \label{fig:Process_Phase}
\end{figure}

\autoref{fig:Process_Phase} shows the progression of the time series recorded with the indirect force sensor. The force curve characterises the process cycle during a press stroke. The measurement is started by the trigger which is activated by the ram moving downwards. The downholer plates touch down at point A and press the strip material onto the die. Between point A and point B, the downholder springs are compressed, causing the applied force to increase linearly. The deep drawing process begins at point B. At point C, the press reaches its bottom dead centre and the reverse stroke begins so that the punch moves out of the material again. At point D, the deep-drawing punch is released from the material and now the hold-down springs relax linearly up to point E. At point E, the downholder plate lifts off again and the component is fed to the last process step.

\begin{table}
\centering
\small
\caption{Overview of conducted runs on the high-speed press with normal and defect states at different stroke rates.}
\label{tab:me_experiments}
\vskip 0.1in
\begin{tabular}{clcc}
\toprule
Experiment \# & State  & Stroke Rate & Samples \\ \midrule
1       & Normal & 80   &  193 \\
2       & Normal & 100  &  193 \\
3       & Normal & 150  &  189 \\ 
4       & Normal & 175  &  198 \\
5       & Normal & 200  &  194 \\
6       & Normal & 225  &  188 \\ 
7       & Defect & 80   &  149 \\
8       & Defect & 100  &  193 \\
9       & Defect & 150  &  188 \\ 
10      & Defect & 175  &  196 \\
11      & Defect & 200  &  193 \\
12      & Defect & 225  &  190 \\ 
\midrule
Total   &        &      &  2264\\         
\bottomrule
\end{tabular}
\end{table}

\subsection{Shortcuts}

In \data, the operation speed of the progressive tool acts as a shortcut to solve the task. The higher the stroke rate of the press, the more friction occurs, and the higher the impact of the downholder plate. The differences can be observed in \autoref{fig:p2s_example}. Since we are aware of these physics-based shortcuts, we are able to annotate them in our dataset. As the process speed increases, the friction that occurs between the die and the material in the deep-drawing stage changes, as the frictional force is dependent on the frictional speed. This is particularly evident in the present case, as there are no deep-drawing oils used in the experiments, which could optimize the friction condition. The areas affected by the punch's friction are from 1380 to 1600 (start of deep drawing) and from 2080 to 2500 (end of deep drawing). In addition, the impulse of the downholder plate affects the die increases due to increased process dynamics. If the process speed is increased, the process force also increases in the ranges of the time series from 800 to 950 (downholder plate sets down) and 3250 to 3550 (downholder plate lifts).

In the experiment setting of \autoref{tab:mechanical_data}, the stroke rate in the training data correlates with the class label, i.e., there are only normal experiments with small stroke rates and defective ones with high stroke rates: Experiment 1, 2, 3, 10, 11, 12 are the training data and the remaining experiments are the test data. To obtain a reference setting without shortcuts, normal and defect experiments are in the same set (training or test). This results in experiments 1, 3, 5, 7, 9, 11 in the training set and the remaining in the test set.

\end{document}